\documentclass[runningheads]{llncs}
\usepackage[T1]{fontenc}
\usepackage{graphicx}
\usepackage{amsmath,amssymb} 
\usepackage{color}

\usepackage{colortbl}
\usepackage{tikz}
\usepackage{comment}
\usepackage{booktabs}
\usepackage{times}
\usepackage{epsfig}
\usepackage{multirow}
\usepackage{enumitem}
\usepackage{bbm}
\usepackage{arydshln}

\usepackage{amsthm}
\usepackage{amsfonts}
\usepackage{dsfont}
\usepackage{bm}
\usepackage{mathtools}
\usepackage{algpseudocode}
\usepackage{algorithm}

\usepackage{subcaption}
\usepackage{xspace}

\usepackage{listings}
\usepackage{color}

\definecolor{dkgreen}{rgb}{0,0.6,0}
\definecolor{gray}{rgb}{0.5,0.5,0.5}
\definecolor{mauve}{rgb}{0.58,0,0.82}

\usepackage[pagebackref,breaklinks,colorlinks,bookmarks=false,urlcolor=black
,citecolor={green!60!black}]{hyperref}

\usepackage{multibib}
\newcites{latex}{References}

\usepackage{filecontents}

\usepackage{orcidlink}

\makeatletter
\renewcommand{\paragraph}{%
  \@startsection{paragraph}{4}%
  {\z@}{0.75ex \@plus 1ex \@minus .2ex}{-0.3em}%
  {\normalfont\normalsize\bfseries}%
}
\makeatother

\renewcommand\vec[1]{\ensuremath\boldsymbol{#1}}
\renewcommand\cdots{...}

\newcommand{\mZ}{\mathbf{Z}}

\newcommand{\vy}{\mathbf{y}}
\newcommand{\valpha}{\bm{\alpha}}

\newcommand{\tD}{\vec{\mathcal{D}}}

\newcommand{\mX}{\mathbf{X}}
\newcommand{\mA}{\mathbf{A}}

\newcommand{\mbrp}[1]{\mathbb{R}_{+}^{#1}}
\newcommand{\mbr}[1]{\mathbb{R}^{#1}}

\newcommand{\tAnb}{\mathcal{A}}

\newcommand{\idx}[1]{\mathcal{I}_{#1}}

\newcommand{\vpsi}{\boldsymbol{\psi}}

\newcommand{\mPsi}{\vec{\Psi}}

\DeclareMathOperator*{\myinf}{Inf}

\DeclareMathOperator*{\softmingg}{SoftMin_{\bar{\gamma}}}
\DeclareMathOperator*{\softming}{SoftMin_\gamma}
\DeclareMathOperator*{\topminb}{TopMin_\beta}
\DeclareMathOperator*{\topmaxbb}{TopMax_{NZ\beta}}

\def\eg{\emph{e.g.}}

\newcommand{\mD}{\boldsymbol{D}}
\newcommand{\vd}{\boldsymbol{d}}

\newcommand{\stkout}[1]{{\ifmmode\text{\sout{\ensuremath{#1}}}\else\sout{#1}\fi}}

\makeatletter
\DeclareRobustCommand\onedot{\futurelet\@let@token\bmv@onedotaux}
\def\bmv@onedotaux{\ifx\@let@token.\else.\null\fi\xspace}
\def\eg{\emph{e.g}\onedot} 
\def\ie{\emph{i.e}\onedot} 
 
\def\etc{\emph{etc}\onedot} \def\vs{\emph{vs}\onedot}
\def\wrt{w.r.t\onedot} 
\def\etal{\emph{et al}\onedot}

\makeatother

\newcommand{\lei}{\textcolor{black}}

\begin{document}
\title{Temporal-Viewpoint Transportation Plan for Skeletal Few-shot Action Recognition}
%
%
\author{Lei Wang\inst{\dagger, \S}\orcidlink{0000-0002-8600-7099} \and
Piotr Koniusz\inst{\S,\dagger}\thanks{This work has been accepted as an oral paper at the 16\textsuperscript{th} Asian Conference on Computer Vision (ACCV'22). It extends our ArXiv 2021 draft JEANIE \cite{jeanie}.}\orcidlink{0000-0002-6340-5289}}
\authorrunning{Wang and Koniusz}
%
\institute{$^{\dagger}$Australian National University \;
   $^\S$Data61/CSIRO\\
   $^\S$firstname.lastname@data61.csiro.au
}
\maketitle


\begin{abstract}
We propose a Few-shot Learning pipeline for 3D skeleton-based action recognition by Joint tEmporal and cAmera viewpoiNt alIgnmEnt (JEANIE). To factor out misalignment between query and support sequences of 3D body joints, we propose an advanced variant of Dynamic Time Warping which jointly models each smooth path between the query and support frames to achieve simultaneously the best alignment in the temporal and simulated camera viewpoint spaces for end-to-end learning under the limited few-shot training data. Sequences are encoded with a temporal block encoder based on Simple Spectral Graph Convolution, a lightweight linear Graph Neural Network backbone. We also include a setting with a transformer.  Finally, we propose a similarity-based loss which encourages the alignment of sequences of the same class while preventing the alignment of unrelated sequences. We show state-of-the-art results on NTU-60, NTU-120, Kinetics-skeleton and UWA3D Multiview Activity II.
\end{abstract}

\section{Introduction}
\label{sec:intro}
Action recognition 
is arguably among key topics in computer vision due to  applications in video surveillance~\cite{lei_thesis_2017,lei_icip_2019}, human-computer interaction, sports analysis, virtual reality and robotics. Many pipelines~\cite{Tran_2015_ICCV,Feichtenhofer_2016_CVPR,Feichtenhofer_2017_CVPR,Carreira_2017_CVPR,lei_tip_2019,hosvd}   perform action classification given the large amount of labeled training data. However, manually collecting and labeling videos for 3D skeleton sequences is laborious, and such pipelines need to be retrained or fine-tuned for new class concepts. Popular  action recognition networks include two-stream neural networks~\cite{Feichtenhofer_2016_CVPR,Feichtenhofer_2017_CVPR,Wang_2017_CVPR} and 3D convolutional networks (3D CNNs)~\cite{Tran_2015_ICCV,Carreira_2017_CVPR}, which  aggregate  frame-wise and temporal block representations, respectively. However, such networks indeed must be  trained on large-scale datasets such as Kinetics \cite{Carreira_2017_CVPR,lei_iccv_2019,lei_mm_21,koniusz2021high} under a fixed set of training class concepts. 

Thus, there exists a growing interest in devising effective  Few-shot Learning (FSL) for action recognition, termed Few-shot Action Recognition (FSAR), that rapidly adapts to  novel classes given a few training samples~\cite{mishra2018generative,xu2018dense,guo2018neural,dwivedi2019protogan,hongguang2020eccv,kaidi2020cvpr,udtw_eccv22}. However, FSAR  for videos is scarce due to the volumetric nature of videos and large intra-class variations. 

FSL for image recognition has been widely studied  \cite{miller_one_example,Li9596,NIPS2004_2576,BartU05,fei2006one,lake_oneshot} including contemporary CNN-based FSL methods  
\cite{meta25,f4Matching,f1,f5Model-Agnostic,f8Relation,sosn}, which use meta-learning, prototype-based learning and feature representation learning. Just in 2020--2022, many FSL methods \cite{guo2020broader,nikita2020eccv,shuo2020eccv,moshe2020eccv,qinxuan2021wacv,nanyi2020accv,jiechao2020accv,kai2020cvpr,thomas2020cvpr,kaidi2020cvpr,luming2020cvpr,maxexp,christian_modgrad,mlso,hao_fsl,keypoint_fsl} have been dedicated to image classification or detection \cite{xin6dof,Zhang_2021_CVPR,shan_fsl,Zhang_2022_CVPR,Zhang_eccv_2022}. Noteworthy mentioning is the incremental learning paradigm that can also tackle novel classes \cite{incremental}. In this paper, we aim at advancing few-shot recognition of articulated set of connected 3D body joints.


With an exception of very recent models  \cite{liu_fsl_cvpr_2017,Liu_2019_NTURGBD120,2021dml,memmesheimer2021skeletondml,udtw_eccv22,qin_tnnls_22}, FSAR approaches that learn from skeleton-based 3D body joints are  scarce. The above situation prevails despite action recognition from articulated sets of connected body joints, expressed as 3D coordinates, does offer a number of advantages over videos such as (i) the lack of the background clutter, (ii) the volume of data being several orders of magnitude smaller, and (iii) the 3D geometric manipulations of sequences being relatively friendly. 

\begin{figure*}[t]
\includegraphics[width=\linewidth]{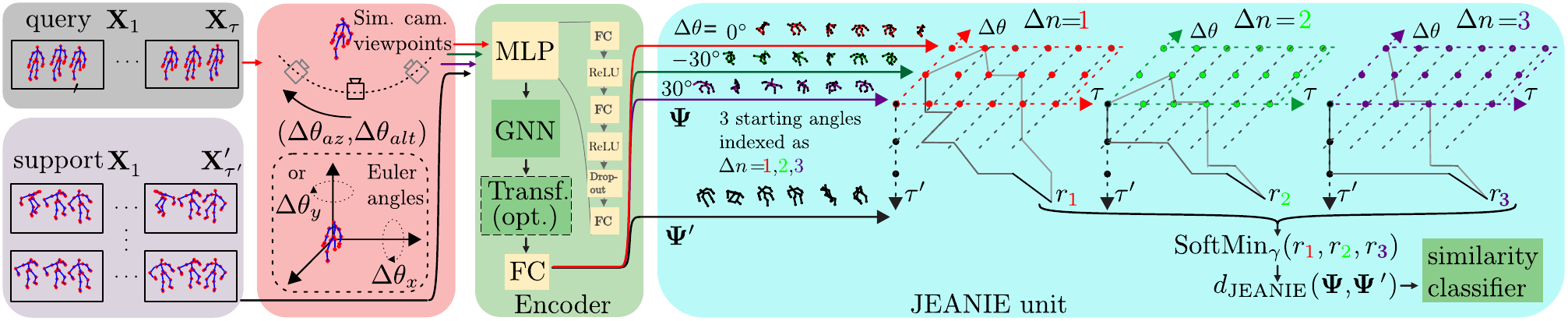}
\caption{Our 3D skeleton-based FSAR with JEANIE. Frames from a query sequence and a support sequence are split into short-term temporal blocks $\mX_1,\cdots,\mX_{\tau}$ and $\mX'_1,\cdots,\mX'_{\tau'}$ of length $M$ given stride $S$. Subsequently, we generate (i) multiple rotations by $(\Delta\theta_x,\Delta\theta_y)$ of each query skeleton by either Euler angles (baseline approach) or (ii) simulated camera views (gray cameras) by camera shifts  $(\Delta\theta_{az},\Delta\theta_{alt})$  \wrt the assumed average camera location (black camera). We pass all skeletons via Encoding Network (with an optional transformer) to obtain feature tensors $\mPsi$ and $\mPsi'$, which are directed to JEANIE. We note that the temporal-viewpoint alignment takes place in 4D space (we show a 3D case with three views: $-30^\circ, 0^\circ, 30^\circ$). Temporally-wise, JEANIE starts from the same $t\!=\!(1,1)$ and finishes at $t\!=\!(\tau,\tau')$ (as in DTW). Viewpoint-wise, JEANIE starts from every possible camera shift $\Delta\theta\in\{-30^\circ, 0^\circ, 30^\circ\}$ (we do not know the true correct pose) and finishes at one of possible camera shifts. At each step, the path may move by no more than $(\pm\!\Delta\theta_{az},\pm\!\Delta\theta_{alt})$ to prevent erroneous alignments. Finally, SoftMin picks up the smallest distance.}
\label{fig:pipe}
\end{figure*}

Thus, we propose a  FSAR  approach that learns on   skeleton-based 3D body joints via Joint tEmporal and cAmera  viewpoiNt alIgnmEnt (JEANIE). As FSL is based on learning similarity between support-query pairs, to achieve good matching of queries with support sequences representing the same action class, we propose to simultaneously model the optimal (i) temporal and (ii) viewpoint alignments. To this end, we build on  soft-DTW \cite{marco2017icml}, a differentiable variant of Dynamic Time Warping (DTW) \cite{marco2011icml}. Unlike soft-DTW, we exploit the projective camera geometry.  We assume that the best smooth path in DTW should simultaneously provide the best temporal and viewpoint alignment, as sequences that are being matched might have been captured under different camera viewpoints or subjects might have followed different trajectories. 

To obtain skeletons under several viewpoints, we  rotate  skeletons (zero-centered by hip) by Euler angles \cite{eulera} \wrt $x$, $y$ and $z$ axes, or generate skeleton locations given simulated camera positions, according to the algebra of stereo projections \cite{sterproj}.

We note that  view-adaptive models 
for action recognition do exist. View Adaptive Recurrent Neural Networks \cite{Zhang_2017_ICCV,8630687} is a classification model equipped with a view-adaptive subnetwork that contains the rotation and translation switches within its RNN backbone, and the main LSTM-based network. Temporal Segment Network \cite{wang_2019_tpami} models long-range temporal structures with a new segment-based sampling and aggregation module. However, such   pipelines  require a large number of training samples with varying viewpoints and temporal shifts  to learn a robust model. Their limitations become evident when a network trained under a  fixed set of action classes has to be adapted to samples of novel classes. 
Our JEANIE does not suffer from such a limitation. 

Our pipeline consists of an MLP  which takes neighboring frames to form a temporal block. Firstly, we sample desired Euler rotations or simulated camera viewpoints, generate multiple skeleton views, and pass them to the MLP to get block-wise feature maps, next forwarded to a Graph Neural Network (GNN), \eg, GCN~\cite{kipf2017semi}, Fisher-Bures GCN \cite{uai_ke}, SGC~\cite{felix2019icml}, APPNP~\cite{johannes2019iclr} or S$^2$GC~\cite{hao2021iclr,coles_hao}, followed by an optional transformer~\cite{dosovitskiy2020image}, and an FC layer to obtain graph-based  representations passed to  JEANIE. 

JEANIE builds on Reproducing Kernel Hilbert Spaces (RKHS) \cite{Smola03kernelsand} which scale gracefully to FSAR problems which, by their setting, learn to match pairs of sequences  rather than predict class labels. JEANIE builds on  Optimal Transport \cite{ot} by using a transportation plan for temporal and viewpoint alignment in skeletal action recognition. 

\vspace{0.1cm}
Below are our contributions:
\renewcommand{\labelenumi}{\roman{enumi}.}
\begin{enumerate}[leftmargin=0.6cm]
\item We propose a Few-shot Action Recognition approach for learning on skeleton-based articulated 3D body joints via JEANIE, which performs the joint alignment of temporal blocks  and simulated viewpoint indexes of skeletons between support-query sequences to select the smoothest path without abrupt jumps in matching temporal locations and view indexes. Warping jointly temporal locations and simulated viewpoint indexes helps meta-learning with limited samples of novel classes.
\item To simulate different viewpoints of 
3D  skeleton  sequences, we consider rotating them  (1) by Euler angles within a specified range along $x$ and $y$ axes, or (2) towards the  simulated camera locations based on the  algebra of stereo projection. 
\item We investigate several different GNN backbones (including transformer), as well as the optimal temporal size and stride for temporal blocks encoded by a simple 3-layer MLP unit before forwarding them to GNN.
\item We propose a simple  similarity-based loss encouraging the alignment of within-class sequences and preventing the alignment of between-class sequences. 
\end{enumerate}

We achieve the state of the art  on 
 large-scale NTU-60 \cite{Shahroudy_2016_NTURGBD},   NTU-120 \cite{Liu_2019_NTURGBD120}, Kinetics-skeleton~\cite{stgcn2018aaai} and  UWA3D Multiview Activity II \cite{Rahmani2016}.
As far as we can tell, the simultaneous alignment in the joint temporal-viewpoint space for FSAR is a novel proposition.

\section{Related Works}
\label{sec:rel}
Below, we describe 3D skeleton-based  action recognition, FSAR approaches and GNNs. 



\paragraph{Action recognition (3D skeletons).}
%
  3D skeleton-based  action recognition pipelines often 
 use GCNs \cite{kipf2017semi}, \eg,  
spatio-temporal GCN \cite{stgcn2018aaai},  
an  a-links inference model \cite{Li_2019_CVPR},  shift-graph model 
\cite{Cheng_2020_CVPR} and multi-scale aggregation node 
\cite{Liu_2020_CVPR}. However, such models  
rely on large-scale datasets, and cannot be easily adapted to novel class concepts. 

\paragraph{FSAR (videos).} 
Approaches \cite{mishra2018generative,guo2018neural,xu2018dense} use a generative model, graph matching on 3D coordinates and dilated networks, 
respectively. Approach \cite{Zhu_2018_ECCV} uses a  compound memory network. 
ProtoGAN \cite{dwivedi2019protogan}  generates action prototypes. Model \cite{hongguang2020eccv} uses permutation-invariant attention and second-order aggregation of temporal video blocks, whereas approach   \cite{kaidi2020cvpr} proposes a modified temporal alignment for query-support pairs via DTW. 

\paragraph{FSAR (3D skeletons).} 
Few FSAR models use 3D skeletons 
\cite{liu_fsl_cvpr_2017,Liu_2019_NTURGBD120,2021dml,memmesheimer2021skeletondml}. Global Con-text-Aware Attention LSTM \cite{liu_fsl_cvpr_2017} selectively focuses on informative joints. 
Action-Part Semantic Relevance-aware (APSR) model \cite{Liu_2019_NTURGBD120}  uses  the  semantic relevance between each body part and  action class at the  distributed  word  embedding  level. 
Signal Level Deep Metric Learning (DML) \cite{2021dml}  and Skeleton-DML \cite{memmesheimer2021skeletondml} one-shot FSL approaches  encode  signals  into  images,  extract  features  using  CNN and apply multi-similarity miner losses. 
In contrast, we use temporal blocks of  3D body joints of skeletons encoded by GNNs under multiple viewpoints of skeletons to simultaneously perform temporal and viewpoint-wise  alignment of query-support in the meta-learning regime.

\paragraph{Graph Neural Networks.} GNNs are popular in the skeleton-based action recognition. 
We build on GNNs in this paper due to their excellent ability to represent graph-structured data such as interconnected body joints. GCN \cite{kipf2017semi} applies graph convolution in the spectral domain, and enjoys the depth-efficiency when stacking multiple layers due to non-linearities. However, depth-efficiency costs speed due to backpropagation through consecutive layers. In contrast, a very recent family of so-called spectral filters do not require depth-efficiency but apply filters based on heat  diffusion to the graph Laplacian. As a result, they are fast linear models as learnable weights act on filtered node representations. SGC~\cite{felix2019icml}, APPNP~\cite{johannes2019iclr} and S$^2$GC~\cite{hao2021iclr} are  three methods from this family which we investigate for the backbone. 

\paragraph{Multi-view action recognition.} 
Multi-modal sensors enable multi-view action recognition \cite{lei_tip_2019,Zhang_2017_ICCV}. 
A Generative Multi-View Action Recognition framework  \cite{Wang_2019_ICCV}  integrates complementary information from  RGB and depth sensors by View Correlation Discovery Network. Some works exploit multiple views of the  subject  \cite{Shahroudy_2016_NTURGBD,Liu_2019_NTURGBD120,8630687,Wang_2019_ICCV} to overcome the viewpoint variations for action recognition on large training datasets. In contrast, our  JEANIE learns to perform jointly the  temporal and simulated viewpoint alignment in an end-to-end meta-learning setting. This is a novel paradigm  based on similarity learning of support-query pairs rather than learning class concepts. 

\section{Approach}
\label{sec:appr}

\begin{figure*}[t]
\begin{minipage}{0.49\linewidth}
\centering
\includegraphics[trim=0 0 0 0, clip=true,width=0.86\linewidth]{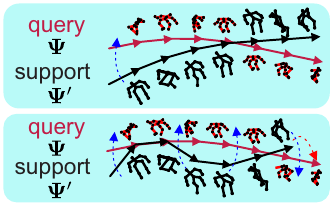}
\caption{\label{fig:sq-match}({\em top}) In viewpoint-invariant learning, the distance between query features $\mPsi$ and support features $\mPsi'$ has to be computed. The blue arrow indicates that trajectories of both actions need alignment. ({\em bottom}) In real life, subject's 3D body joints deviate from one ideal trajectory, and so advanced viewpoint alignment strategy is needed.}
%

\end{minipage}
$\;\;\;$
\begin{minipage}{0.49\linewidth}
\centering
\includegraphics[trim=0cm 0cm 0cm 0cm, clip=true,width=0.86\linewidth]{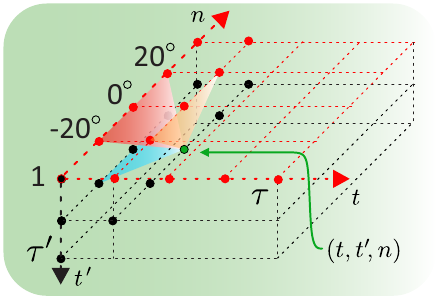}
\caption{\label{fig:jeanie_temp-shift}JEANIE (1-max shift). 
We loop over all points. At $(t,t',n)$ (green point) we add its base distance to the minimum of accumulated distances at $(t,t'\!\!-\!1,n\!-\!1)$, $(t,t'\!\!-\!1,n)$, $(t,t'\!\!-\!1,n\!+\!1)$ (orange plane), $(t\!-\!1,t'\!\!-\!1,n\!-\!1)$, $(t\!-\!1,t'\!\!-\!1,n)$, $(t\!-\!1,t'\!\!-\!1,n\!+\!1)$ (red plane) and  $(t\!-\!1,t'\!,n\!-\!1)$, $(t\!-\!1,t'\!,n)$, $(t\!-\!1,t'\!,n\!+\!1)$ (blue plane). 
}
\end{minipage}
\end{figure*}

\begin{figure*}[t]
\centering
%

\begin{subfigure}[t]{0.32\linewidth}
\centering\includegraphics[trim=2.0cm 2.25cm 0.5cm 3.5cm, clip=true, width=\linewidth]{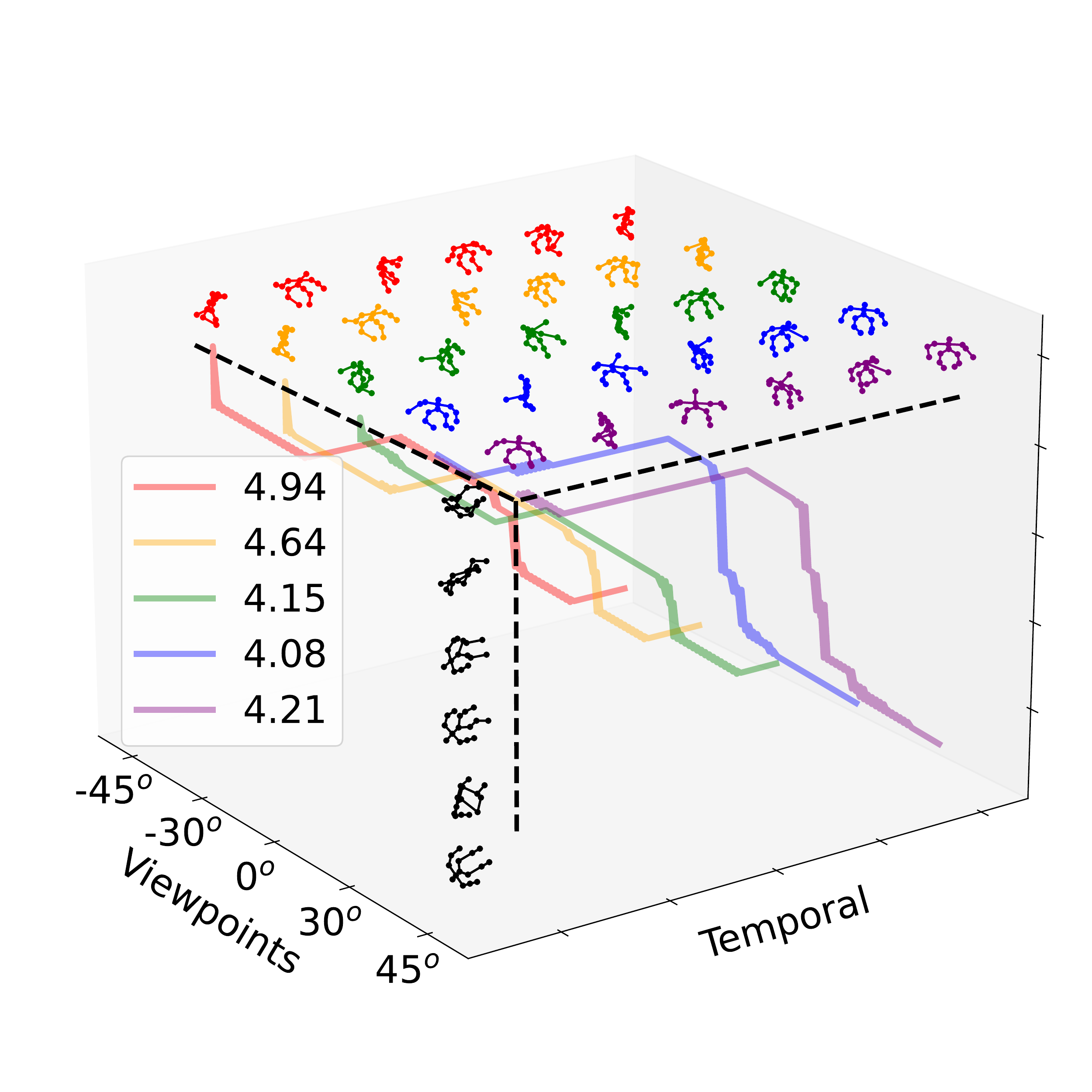}
\caption{soft-DTW (view-wise)}\label{fig:sdtw}
\end{subfigure}
\begin{subfigure}[t]{0.32\linewidth}
\centering\includegraphics[trim=2.0cm 2.25cm 0.5cm 3.5cm, clip=true, width=\linewidth]{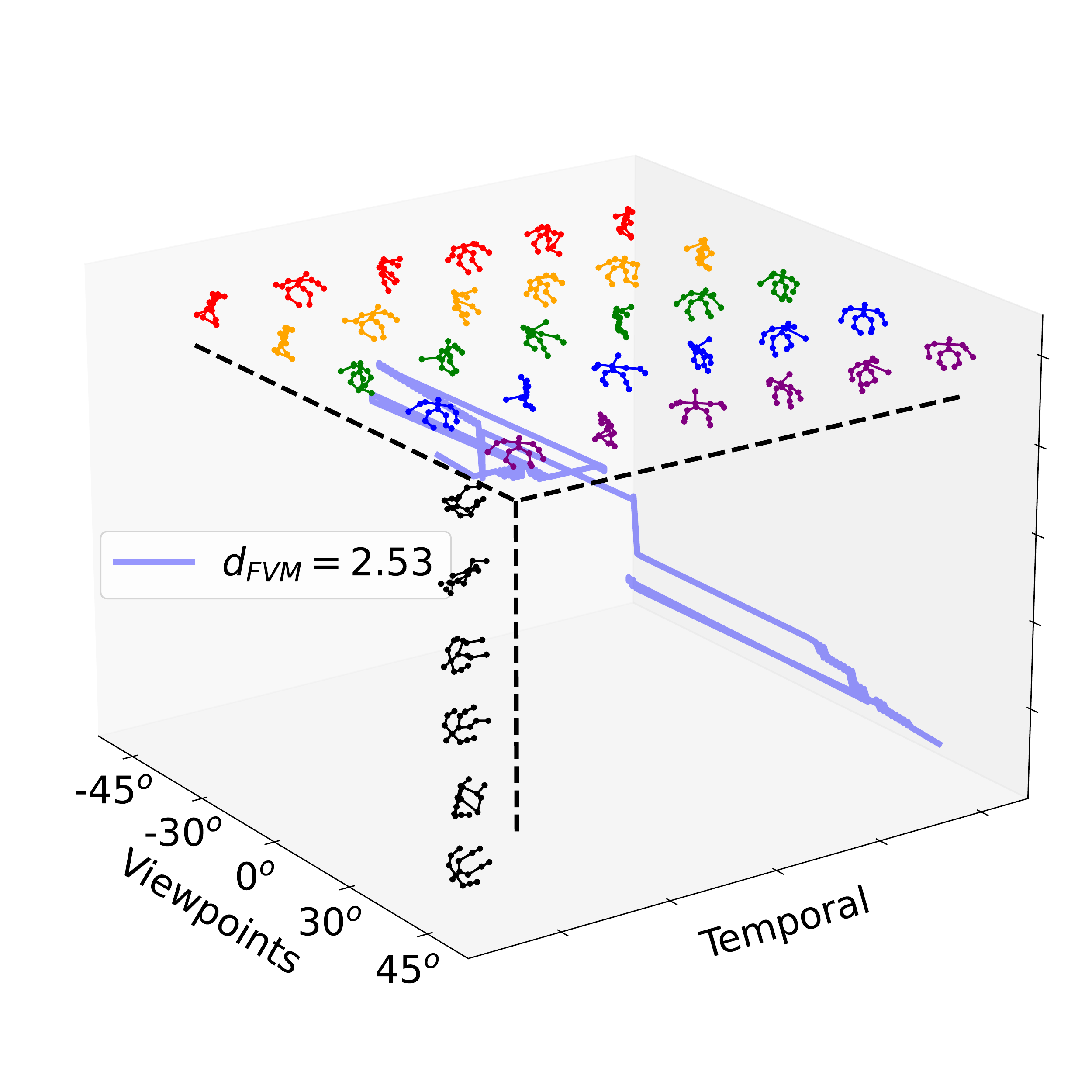}
\caption{FVM}\label{fig:fvm}
\end{subfigure}
\begin{subfigure}[t]{0.32\linewidth}
\centering\includegraphics[trim=2.0cm 2.25cm 0.5cm 3.5cm, clip=true, width=\linewidth]{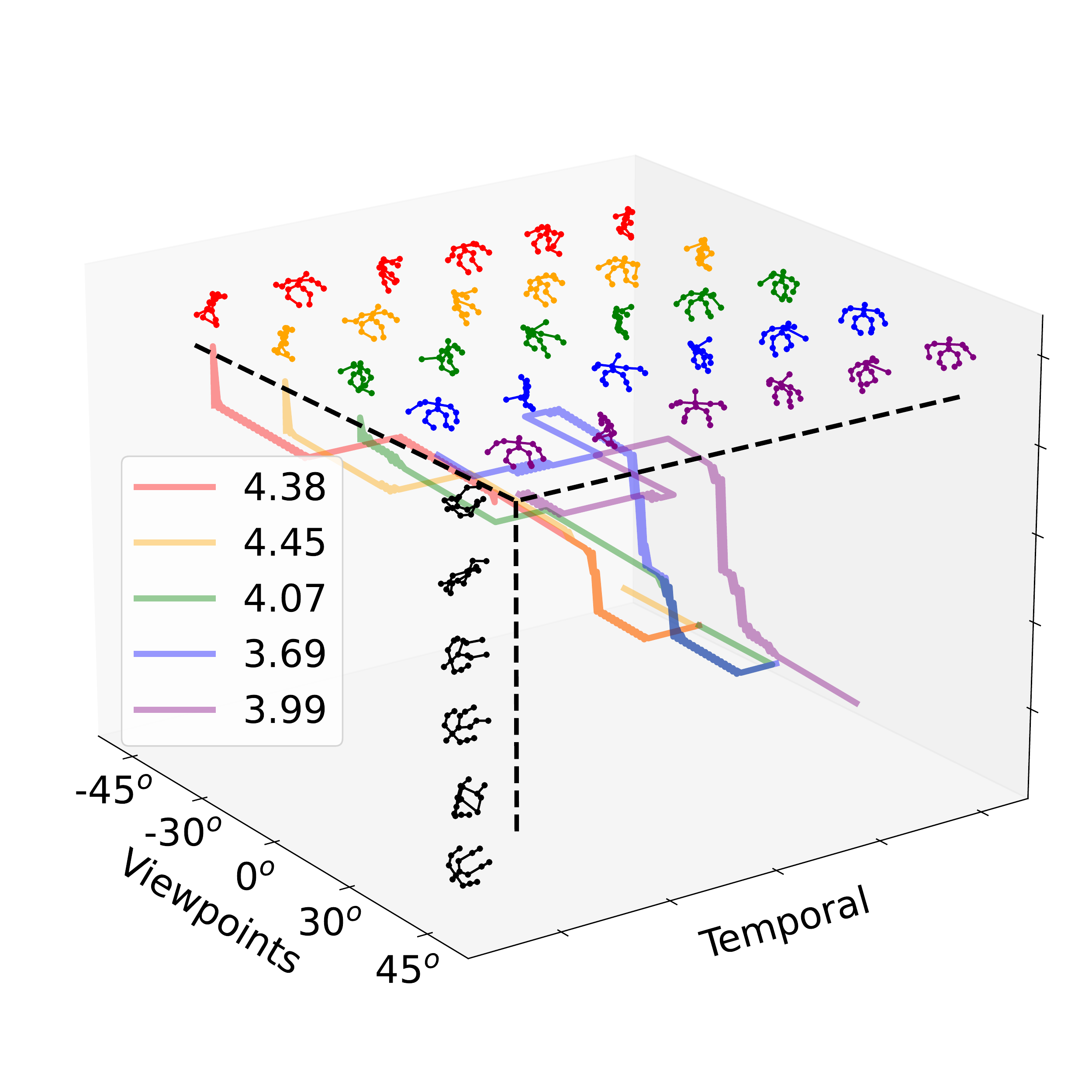}
\caption{JEANIE (1-max shift)}\label{fig:jeanie-v2}
\end{subfigure}
%

%
%
\caption{\lei{A comparison of paths in 3D for soft-DTW, Free Viewpoint Matching (FVM) and our JEANIE.
For a given support skeleton sequence (green color), we choose viewing angles between $-45^\circ$ and $45^\circ$ for the camera viewpoint simulation. The support skeleton sequence is shown in black color.
(a) soft-DTW finds each individual alignment per viewpoint fixed throughout alignment:  $d_\text{shortest}\!=\!4.08$. (b) FVM is a greedy matching algorithm that in each time step  seeks the best alignment pose from all viewpoints which leads to unrealistic zigzag path (person cannot jump from front to back view suddenly): $d_\text{FVM}\!=\!2.53$. (c) Our JEANIE (1-max shift) is able to find smooth joint viewpoint-temporal alignment between support and query sequences. We show each optimal path for each possible starting position: $d_\text{JEANIE}\!=\!3.69$. While $d_\text{FVM}\!=\!2.53$ for FVM is overoptimistic, $d_\text{shortest}\!=\!4.08$ for fixed-view matching is too pessimistic, whereas JEANIE strikes the right matching balance with $d_\text{JEANIE}\!=\!3.69$.}}
\label{fig:jeanie_fvm_plots}
\end{figure*}

To learn similarity/dissimilarity between pairs of sequences of 3D body joints representing query and support samples from episodes, our goal is to find a smooth joint viewpoint-temporal alignment of query and support and minimize or maximize the matching distance $d_\text{JEANIE}$ (end-to-end setting) for same or different support-query labels, respectively. 
Fig.~\ref{fig:sq-match} (top) shows that sometimes matching of query and support may be as easy as rotating one trajectory onto another, in order to achieve viewpoint invariance. 
A viewpoint invariant distance \cite{inv_kern} can be defined as:
\begin{equation}
d_{\text{inv}}(\mPsi, \mPsi')\!=\!\myinf\limits_{\gamma,\gamma'\in T} d\big(\gamma(\mPsi), \gamma'(\mPsi')\big),
\label{eq:inv}
\end{equation}
where $T$ is a set of transformations required to achieve a viewpoint invariance, $d(\cdot,\cdot)$ is some base distance, \eg, the Euclidean distance, and $\mPsi$ and $\mPsi'$ are features describing query and support pair of sequences. Typically, $T$ may include 3D rotations to rotate one trajectory onto the other. However, such a global viewpoint alignment of two sequences is suboptimal. Trajectories are unlikely to be straight 2D lines in the 3D space. Fig.~\ref{fig:sq-match} (bottom) shows that 3D body joints locally follow complicated non-linear paths.

%
%
%
Thus, we propose JEANIE that aligns and warps query/support sequences based on the feature similarity. One can think of JEANIE as performing Eq. \eqref{eq:inv} with $T$ containing camera viewpoint rotations, and the base distance $d(\cdot,\cdot)$ being a joint temporal-viewpoint variant of soft-DTW to account for local temporal-viewpoint variations of 3D body joint trajectories. JEANIE unit in Fig. \ref{fig:pipe} realizes such a strategy (SoftMin operation is equivalent of Eq. \eqref{eq:inv}). While such an idea sounds simple, it is effective, it has not been done before. Fig. \ref{fig:jeanie_temp-shift} (discussed later in the text) shows one step of the temporal-viewpoint computations of JEANIE.

We present a necessary background on Euler angles and the algebra of stereo projection, GNNs and the formulation of soft-DTW in Appendix Sec.~\ref{supp:prereq}. Below, we detail our pipeline shown in Figure \ref{fig:pipe}, explain the proposed JEANIE and our loss function. 

\paragraph{Notations.}  \lei{$\idx{K}$ stands for the index set $\{1,2,\cdots,K\}$. Concatenation of $\alpha_i$ is denoted by $[\alpha_i]_{i\in\idx{I}}$, whereas $\mX_{:,i}$ means we extract/access column $i$ of matrix $\mD$. Calligraphic mathcal fonts denote tensors (\eg, $\tD$), capitalized bold symbols are matrices (\eg, $\mD$), lowercase bold symbols  are vectors (\eg, $\vpsi$), and regular fonts denote scalars.}

\paragraph{Encoding Network (EN).}  
We start by generating $K\!\times\!K'$ Euler rotations or $K\!\times\!K'$ simulated camera views (moved gradually from the estimated camera location) of query skeletons. 
Our EN contains a simple 3-layer MLP unit (FC, ReLU, FC, ReLU, Dropout, FC), GNN, optional  Transformer~\cite{dosovitskiy2020image} and FC. 
The MLP unit takes $M$ neighboring frames, each with $J$ 3D skeleton body joints, forming one temporal block. In total, depending on stride $S$, we obtain some $\tau$ temporal blocks which capture the short temporal dependency, whereas the long temporal dependency is modeled with our JEANIE. Each temporal block is encoded by the MLP into a $d\!\times\!J$ dimensional feature map.  Subsequently,  query feature maps of size $K\!\times\!K'\!\times\!\tau$ and support feature maps of size $\tau'$  are  forwarded to a GNN, optional Transformer (similar to ViT~\cite{dosovitskiy2020image}, instead of using image patches, we feed each body joint encoded by GNN into the transformer), and  an FC layer, which returns $\mPsi\!\in\!\mbr{d'\times K\times K'\times\tau}$ query feature maps and $\mPsi'\!\in\!\mbr{d'\times\tau'}\!$ support feature maps.
Feature maps are  passed to JEANIE and the similarity classifier. 

Let support  maps  $\mPsi'\!$ be 
$[f(\boldsymbol{X}'_1;\mathcal{F}),\cdots,f(\boldsymbol{X}'_{\tau'};\mathcal{F})]\!\in\!\mbr{d'\times\tau'}$
and query maps $\mPsi$ be 
$[f(\boldsymbol{X}_1;\mathcal{F}),\cdots,f(\boldsymbol{X}_\tau;\mathcal{F})]\!\in\!\mbr{d'\times K\times K'\times\tau}$, for  
query and support frames per block $\mX,\mX'\!\in\!\mbr{3\times J\times M}$. Moreover, we define {\fontsize{9}{9}\selectfont$f(\mX; \mathcal{F})\!=\!\text{FC}(\text{Transf}(\text{GNN}(\text{MLP}(\mX; \mathcal{F}_{MLP}); $ $\mathcal{F}_{GNN}); \mathcal{F}_{Transf}); \mathcal{F}_{FC})$},  $\mathcal{F}\!\equiv\![\mathcal{F}_{MLP},\mathcal{F}_{GNN},\mathcal{F}_{Transf},\mathcal{F}_{FC}]$ is the  set of  parameters of EN (note optional Transformer \cite{dosovitskiy2020image}). As GNN, we try GCN~\cite{kipf2017semi}, SGC~\cite{felix2019icml}, APPNP~\cite{johannes2019iclr} or S$^2$GC~\cite{hao2021iclr}.

\paragraph{JEANIE.}  Matching query-support pairs requires temporal alignment due to potential offset in locations of discriminative parts of actions, and due to potentially different dynamics/speed of actions taking place. The same concerns the direction of the dominant action trajectory \wrt the camera. Thus, JEANIE, our advanced soft-DTW, has the  transportation plan  $\tAnb'\!\equiv\!\tAnb_{\tau,\tau', K, K'}$, where apart from temporal block counts $\tau$ and $\tau'$, for query sequences, we have possible $\eta_{az}$ left and $\eta_{az}$ right steps from the initial camera azimuth, and $\eta_{alt}$ up and $\eta_{alt}$ down  steps from the initial camera altitude. Thus, $K\!=\!2\eta_{az}\!+\!1$, $K'\!=\!2\eta_{alt}\!+\!1$.  For the variant with Euler angles, we simply have $\tAnb''\!\equiv\!\tAnb_{\tau,\tau', K, K'}$ where $K\!=\!2\eta_{x}\!+\!1$, $K'\!=\!2\eta_{y}\!+\!1$ instead. Then, JEANIE is given as:
%
\begin{equation}
d_{\text{JEANIE}}(\mPsi,\mPsi')\!=\!\softming\limits_{\mA\in\tAnb'}\left\langle\mA,\tD(\mPsi,\mPsi')\right\rangle,
\label{eq:d_jeanie}
\end{equation}
%
%
where  $\tD\!\in\!\mbrp{K\times\!K'\!\times\tau\times\tau'}\!\!\equiv\![d_{\text{base}}(\vpsi_{m,k,k'},\vpsi'_n)]_{\substack{ (m,n)\in\idx{\tau}\!\times\!\idx{\tau'}\\(k,k')\in\idx{K}\!\times\!\idx{K'}}}\!\!\!$
and $\tD$ contains distances.

Figure \ref{fig:jeanie_temp-shift} shows one step of JEANIE (1-max shift). Suppose the given viewing angle set is $\{-40^{\circ}, -20^{\circ}, 0^{\circ}, 20^{\circ}, 40^{\circ}\}$. For 1-max shift, we loop over $(t,t'\!,n)$. At location $(t,t'\!,n)$, we extract the base distance and add it together with the minimum of aggregated distances at the shown 9 predecessor points. We store that total distance at $(t,t'\!,n)$, and we move to the next point. Note that for viewpoint index $n$, we look up $(n\!-\!1,n,n\!+\!1)$. 
%
Extension to the $\iota$-max shift is straightforward.

Algorithm \ref{code:JEANIE} illustrates JEANIE. For brevity, let us tackle the camera viewpoint alignment  in a single space, \eg, for some shifting steps $-\eta,\cdots,\eta$, each with  size  $\Delta\theta_{az}$. The maximum viewpoint change from block to block is $\iota$-max shift (smoothness). As we have no way to know the initial optimal camera shift, we initialize all possible origins of shifts in accumulator $r_{n,1,1}\!=\!d_{\text{base}}(\vpsi_{n,1}, \vpsi'_{1})$ for all $n\!\in\!\{-\eta, \cdots, \eta\}$. Subsequently, a  phase related to soft-DTW (temporal-viewpoint alignment) takes place. Finally, we choose the path with the smallest distance  over all possible 
viewpoint ends by selecting a soft-minimum over  $[r_{n,\tau,\tau'}]_{n\in\{-\eta, \cdots, \eta\}}$. Notice that accumulator $\boldsymbol{\mathcal{R}}\in\mbr{(2\iota+1)\times\tau\times\tau'}$. Moreover, whenever either index $n\!-\!i$, $t\!-\!j$ or $t'\!-\!k$ in $r_{n\!-\!i,t\!-\!j,t'\!-\!k}$ (see algorithm) is out of bounds, we define $r_{n\!-\!i,t\!-\!j,t'\!-\!k}=\infty$.

\paragraph{FVM.}  To ascertain whether JEANIE is better than performing separately the temporal and simulated viewpoint alignments, we introduce a baseline called the Free Viewpoint Matching (FVM). FVM, for every step of DTW, seeks the best local viewpoint alignment, thus realizing non-smooth temporal-viewpoint path in contrast to JEANIE. To this end, we apply DTW in \lei{Eq.~\eqref{eq:d_jeanie}} with the base distance replaced by:
%
\begin{align}
& d_{\text{FVM}(\vpsi_{t},\vpsi'_{t'})}\!=\!\softmingg\limits_{m,n,m',n'\in\{-\eta,\cdots,\eta\}} d_{\text{base}}(\vpsi_{m,n,t},\vpsi'_{m',n',t'}),
\label{eq:suppl1}
\end{align}
where $\mPsi\!\in\!\mbr{d'\times K\times K'\times\tau}$  and $\mPsi'\!\in\!\mbr{d'\times K\times K'\times\tau'}\!$ are query and support feature maps. We abuse  the notation by writing $d_{\text{FVM}(\vpsi_{t},\vpsi'_{t'})}$ as we minimize over viewpoint indexes in Eq. \eqref{eq:suppl1}. We compute  the distance matrix $\mD\!\in\!\mbrp{\tau\times\tau'}\!\!\equiv\![d_{\text{FVM}}(\vpsi_t,\vpsi'_{t'})]_{(t,t')\in\idx{\tau}\times\idx{\tau'}}$.


\lei{Fig.~\ref{fig:jeanie_fvm_plots} shows the comparison between soft-DTW (view-wise), FVM and our JEANIE. FVM is a greedy matching method which leads to complex zigzag path in 3D space (assuming the camera viewpoint single space in $\vpsi_{n,t}$ and no viewpoint in $\vpsi'_{t'}$). Although FVM is able to find the smallest distance path compared to soft-DTW and JEANIE, it suffers from several issues (i) It is unreasonable for poses in a given sequence to match under sudden jumps in viewpoints.  (ii) Suppose the two sequences are from two different classes, FVM still yields the smallest distance (decreased inter-class variance). 
}

\algblock{while}{endwhile}
\algblock[TryCatchFinally]{try}{endtry}
\algcblockdefx[TryCatchFinally]{TryCatchFinally}{catch}{endtry}
	[1]{\textbf{except}#1}{}
\algcblockdefx[TryCatchFinally]{TryCatchFinally}{elsee}{endtry}
	[1]{\textbf{else}#1}{}
\algtext*{endwhile}
\algtext*{endtry}

\algblock{for}{endfor}
\algtext*{endfor}

\algblockdefx{ifff}{endifff}
	[1]{\textbf{if}#1}{}
\algtext*{endifff}

\algblockdefx{elseee}{endelseee}
	[1]{\textbf{else}#1}{}
\algtext*{endelseee}

\begin{algorithm}[tbp!]
\caption{Joint tEmporal and cAmera viewpoiNt alIgnmEnt (JEANIE).}
\label{code:JEANIE}
{\bf Input} (forward pass): $\mPsi, \mPsi'$, $\gamma\!>\!0$, $d_{\text{base}}(\cdot,\cdot)$, $\iota$-max shift.
\begin{algorithmic}[1]
\State{$r_{:,:,:}\!=\!\infty$, $r_{n,1,1}\!=\!d_{\text{base}}(\vpsi_{n,1}, \vpsi'_{1}),\;\forall n\!\in\!\{-\eta, \cdots, \eta\}$}
\State{$\Pi\equiv\{-\iota,\cdots,0,\cdots,\iota\}\times\{(0,1),(1,0),(1,1)\}$}
\for{ $t\!\in\!\idx{\tau}$:}
\for{ $t'\!\in\!\idx{\tau'}$:}
\ifff{ $t\!\neq\!1$ or $t'\!\!\neq\!1$:}
\for{ $n\!\in\!\{-\eta,\cdots,\eta\}$:}
	\State{$r_{n,t,t'}=d_{\text{base}}(\vpsi_{n,t}, \vpsi'_{t'})+
	\softming\Big([r_{n\!-\!i,t\!-\!j,t'\!-\!k}]_{(i,j,k)\in\Pi}\Big)
	$}
\endfor
	\endifff
\endfor
\endfor
\end{algorithmic}
{\bf Output:} $\softming\Big([r_{n,\tau,\tau'}]_{n\in\{-\eta,\cdots,\eta\}}\Big)$
\end{algorithm}

\paragraph{Loss Function.} \sloppy For the $N$-way $Z$-shot problem, 
we have   one query feature map  and $N\!\times\!Z$ support feature maps per episode. We form a mini-batch containing $B$ episodes. 
Thus, we have query feature maps $\{\mPsi_b\}_{b\in\idx{B}}$ and support feature maps $\{\mPsi'_{b,n,z}\}_{b\in\idx{B},n\in\idx{N},z\in\idx{Z}}$. Moreover,  $\mPsi_b$ and $\mPsi'_{b,1,:}$ share the same class, one of $N$ classes drawn per episode, forming the  subset $C^{\ddagger} \equiv \{c_1,\cdots,c_N \} \subset \mathcal{I}_C \equiv \mathcal{C}$. To be precise, labels  $y(\mPsi_b)\!=\!y(\mPsi'_{b,1,z}), \forall b\!\in\!\idx{B}, z\!\in\!\idx{Z}$ while $y(\mPsi_b)\!\neq\!y(\mPsi'_{b,n,z}), \forall b\!\in\!\idx{B},n\!\in\!\idx{N}\!\setminus\!\{1\},  z\!\in\!\idx{Z}$. In most cases, $y(\mPsi_b)\!\neq\!y(\mPsi_{b'})$ if $b\!\neq\!b'$ and $b,b'\!\in\!\idx{B}$. Selection of $C^{\ddagger}$ per episode is  random. For the $N$-way $Z$-shot protocol, we minimize:
\begin{align}
 & l(\vd^{+}\!,\vd^{-})\!=\!\left(\mu(\vd^{+})\!-\!\{\mu(\topminb(\vd^{+}))\}\right)^2\label{eq:pos}\\
 & \qquad\qquad\!+\!\left(\mu(\vd^{-})\!-\!\{\mu(\topmaxbb(\vd^{-}))\}\right)^2\!,\label{eq:neg}\\
 & 
 \text{where}\;\vd^{+}\!=\![d_\text{JEANIE}(\mPsi_{b},\mPsi'_{b,1,z})]_{\substack{b\in\idx{B}\\z\in\idx{Z}}} 
 \text{ and}\;\vd^{-}\!=\![d_\text{JEANIE}(\mPsi_{b},\mPsi'_{b,n,z})]_{\!\!\!\!\!\substack{b\in\idx{B},\\n\in\idx{N}\!\setminus\!\{1\}\!, z\in\idx{Z}}},\nonumber
\end{align}
where $\vd^+$ is a set of within-class distances for the mini-batch of size $B$ given $N$-way $Z$-shot learning protocol. By analogy,  $\vd^-$ is a set of between-class distances. Function $\mu(\cdot)$ is simply the mean over coefficients of the input vector, $\{\cdot\}$ detaches the graph during the backpropagation step, whereas $\topminb(\cdot)$ and $\topmaxbb(\cdot)$ return $\beta$ smallest and $NZ\beta$ largest coefficients from the input vectors, respectively. Thus, Eq. \eqref{eq:pos} promotes the within-class similarity while Eq. \eqref{eq:neg} reduces the between-class similarity. Integer $\beta\!\geq\!0$ controls the focus on difficult examples, \eg, $\beta\!=\!1$  encourages all within-class distances in Eq.  \eqref{eq:pos} to be close to the positive target $\mu(\topminb(\cdot))$, the smallest observed within-class distance in the mini-batch. If $\beta\!>\!1$, this means we relax our positive target.  By analogy, if $\beta\!=\!1$, we encourage all between-class distances in Eq.  \eqref{eq:neg} to approach the negative target $\mu(\topmaxbb(\cdot))$, the average over the largest $NZ$ between-class distances. If $\beta\!>\!1$, the negative target is relaxed.

\section{Experiments}
\label{sec:exper}
We provide network configurations and training details in Appendix Sec.~\ref{network_train}.
Below, we describe the datasets and evaluation protocols on which we  validate our JEANIE.

\paragraph{Datasets.} Appendix Sec. \ref{app:ds} and Table \ref{datasets} contain details of datasets described below.

\renewcommand{\labelenumi}{\roman{enumi}.}
\begin{enumerate}[leftmargin=0.6cm]
\item{{\em UWA3D Multiview Activity II}}~\cite{Rahmani2016} contains 30 actions performed by 9 people in a cluttered environment. In this dataset, the Kinect camera was moved to different positions to capture the actions from 4 different views: front view ($V_1$), left view ($V_2$), right view ($V_3$), and top view ($V_4$). 

\item{{\em NTU RGB+D (NTU-60)}}~\cite{Shahroudy_2016_NTURGBD} contains 56,880 video sequences and over 4 million frames. 
This dataset has variable sequence lengths  and high intra-class variations.

\item{{\em NTU RGB+D 120 (NTU-120)}}~\cite{Liu_2019_NTURGBD120}, an extension of NTU-60, contains 120 action classes (daily/health-related), and 114,480 RGB+D video samples  captured with 106 distinct human subjects from 155 different camera viewpoints. 

\item{{\em Kinetics}}~\cite{kay2017kinetics} is a large-scale collection of 
650,000 video clips that cover 400/600/700 human action classes. 
It includes human-object interactions such as {\it playing instruments}, as well as human-human interactions such as {\it shaking hands} and {\it hugging}. As the Kinetics-400 dataset provides only the raw videos, we follow approach~\cite{stgcn2018aaai} and use the estimated joint locations in the pixel coordinate system as the input to our pipeline. To obtain the joint locations, we first resize all videos to the resolution of 340 $\times$ 256, and convert the frame rate to 30 FPS. Then we use the publicly available {\it OpenPose}~\cite{Cao_2017_CVPR} toolbox to estimate the location of 18 joints on every frame of the clips. As OpenPose  produces the 2D body joint coordinates  
and Kinetics-400 does not offer multiview or depth data, we use a network of Martinez \etal   \cite{martinez_2d23d} pre-trained on  
Human3.6M~\cite{Catalin2014Human3}, combined with the 2D OpenPose output to  estimate 3D coordinates from 2D coordinates. The 2D OpenPose  and the latter network give us $(x,y)$ and $z$ coordinates, respectively.
\end{enumerate}

\paragraph{Evaluation protocols.} 
For the UWA3D Multiview Activity II, we use standard multi-view classification protocol~\cite{Rahmani2016,lei_thesis_2017,lei_tip_2019}, but we apply it to one-shot learning as the view combinations for training and testing sets are disjoint. For NTU-120, we follow the standard one-shot protocol~\cite{Liu_2019_NTURGBD120}. Based on this protocol, we create a similar one-shot protocol for NTU-60, with  50/10 action classes used for training/testing respectively. To evaluate the effectiveness of the proposed method on viewpoint alignment, we also create two new protocols on NTU-120, for which we group the whole dataset based on (i) horizontal camera views into left, center and right views, (ii) vertical camera views into top, center and bottom views. We conduct two sets of experiments on such disjoint view-wise splits: (i) using 100 action classes for training, and testing on the same 100 action classes (ii) training on 100 action classes but testing on the rest unseen 20 classes. Appendix Sec. \ref{app:epr} details
new/additional eval. protocols on NTU-60/NTU-120. 

\paragraph{Stereo projections.} For simulating different camera viewpoints, we  estimate the fundamental matrix $F$ (Eq.~\eqref{eq:f_matrix} in Appendix),
which
relies on camera parameters. Thus, 
we use the Camera Calibrator from MATLAB to estimate  intrinsic, extrinsic and lens distortion parameters. For a given skeleton dataset, we compute the range of spatial coordinates $x$ and $y$, respectively. We then split them into 3 equally-sized groups to form roughly left, center, right views and other 3 groups for bottom, center, top views. We choose $\sim$15 frame images from each corresponding group, upload them to the Camera Calibrator, and  export  camera parameters.  We then compute the average distance/depth and height per group to estimate the camera position. On NTU-60 and NTU-120, we simply group the whole dataset into 3 cameras, which are left, center and right views, as provided in~\cite{Liu_2019_NTURGBD120}, and then we compute the average distance per camera view based on the height and distance settings given in the table in~\cite{Liu_2019_NTURGBD120}.


\begin{figure}[t]
\begin{minipage}{0.66\linewidth}
\begin{subfigure}[b]{0.49\linewidth}
\includegraphics[trim=0 0 0 0, clip=true,width=\linewidth]{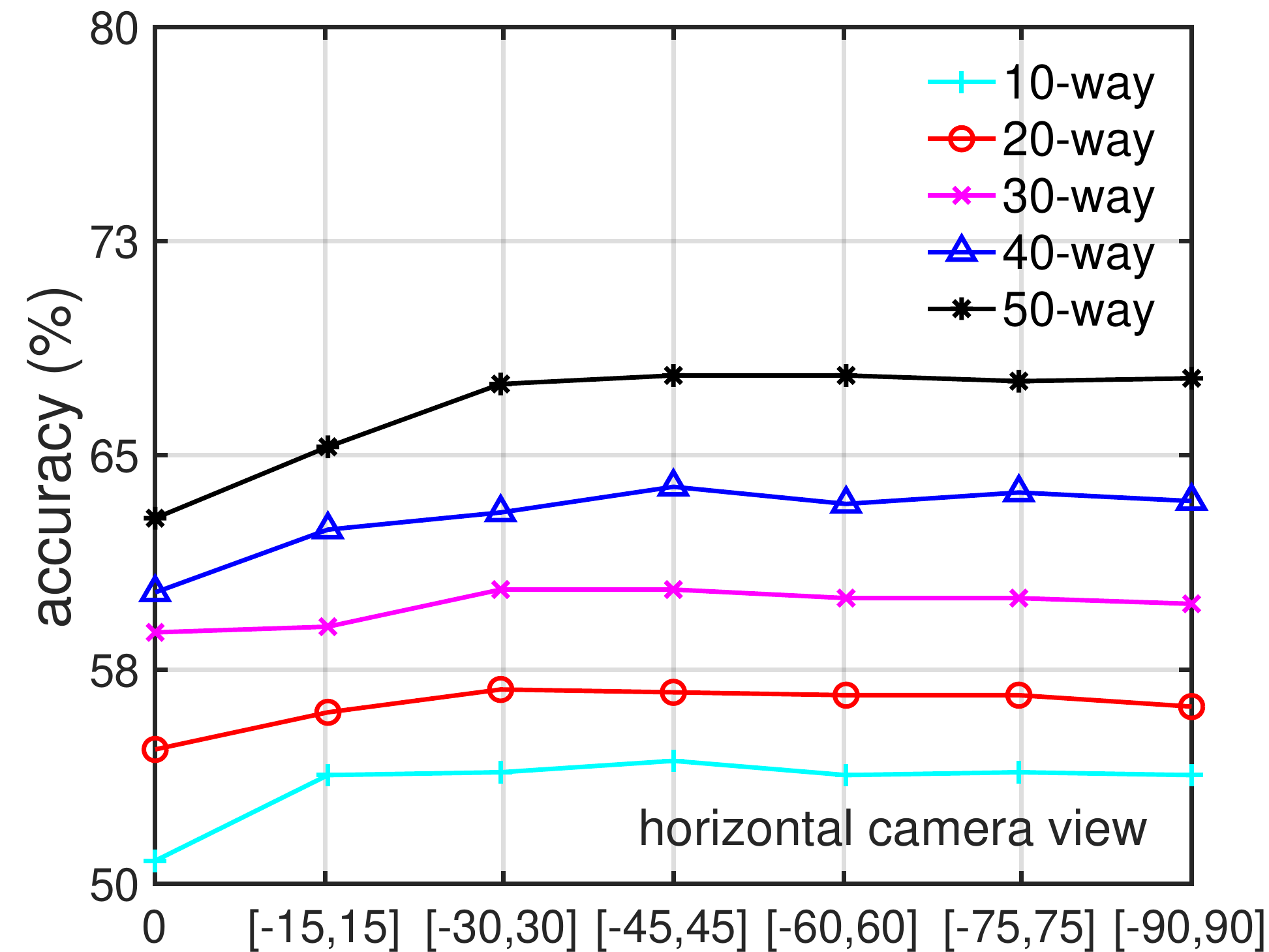}
\caption{\label{fig:sgc_h_angles} horizontal camera view}
\end{subfigure}
\begin{subfigure}[b]{0.49\linewidth}
\includegraphics[trim=0 0 0 0, clip=true,width=\linewidth]{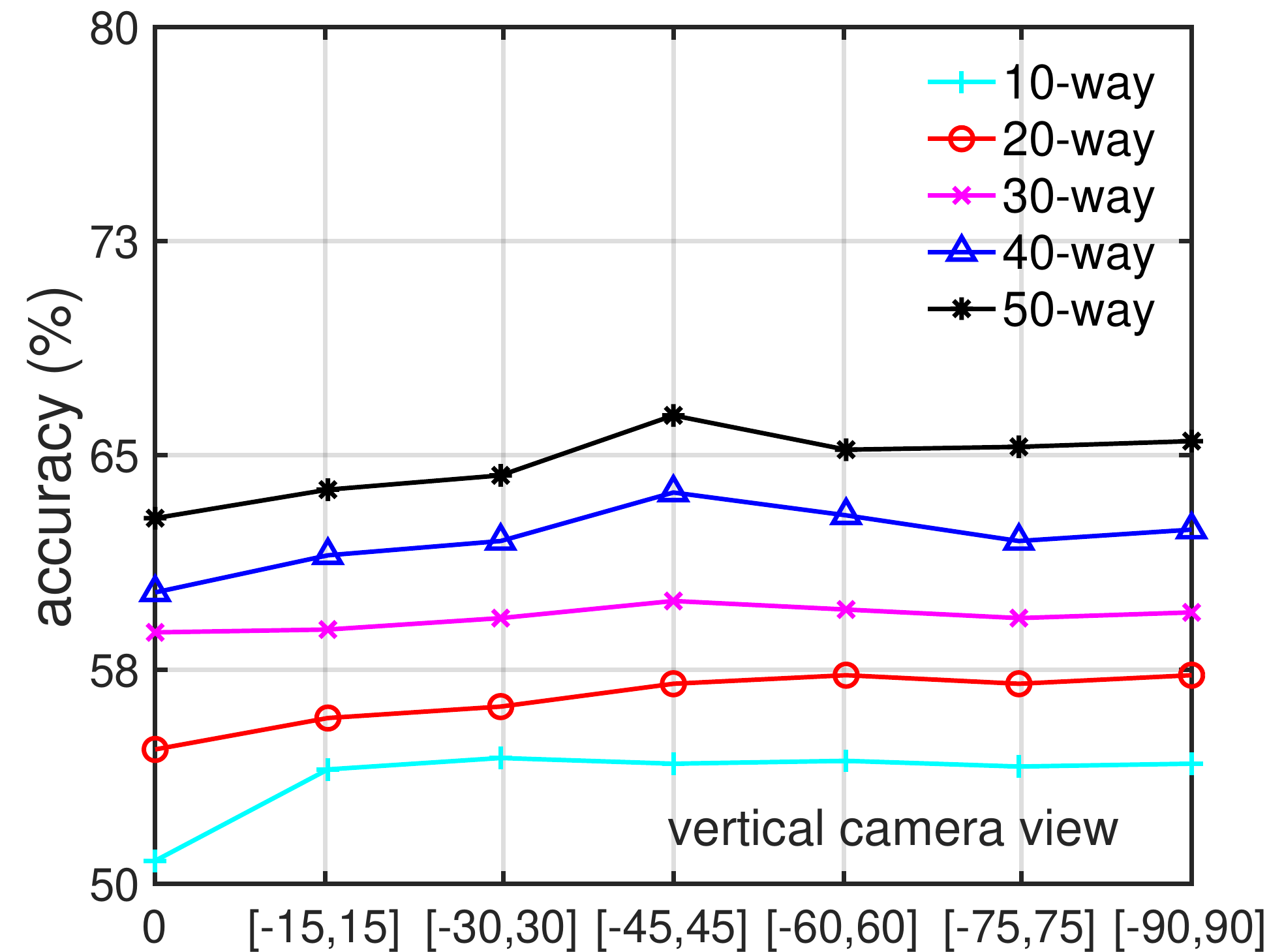}
\caption{\label{fig:sgc_v_angles} vertical camera view}
\end{subfigure}
\caption{The impact of viewing angles on NTU-60.}
\label{fig:h_v_angles}
\end{minipage}
$\;\;\;\;$
\begin{minipage}{0.30\linewidth}
\includegraphics[width=0.95\linewidth]{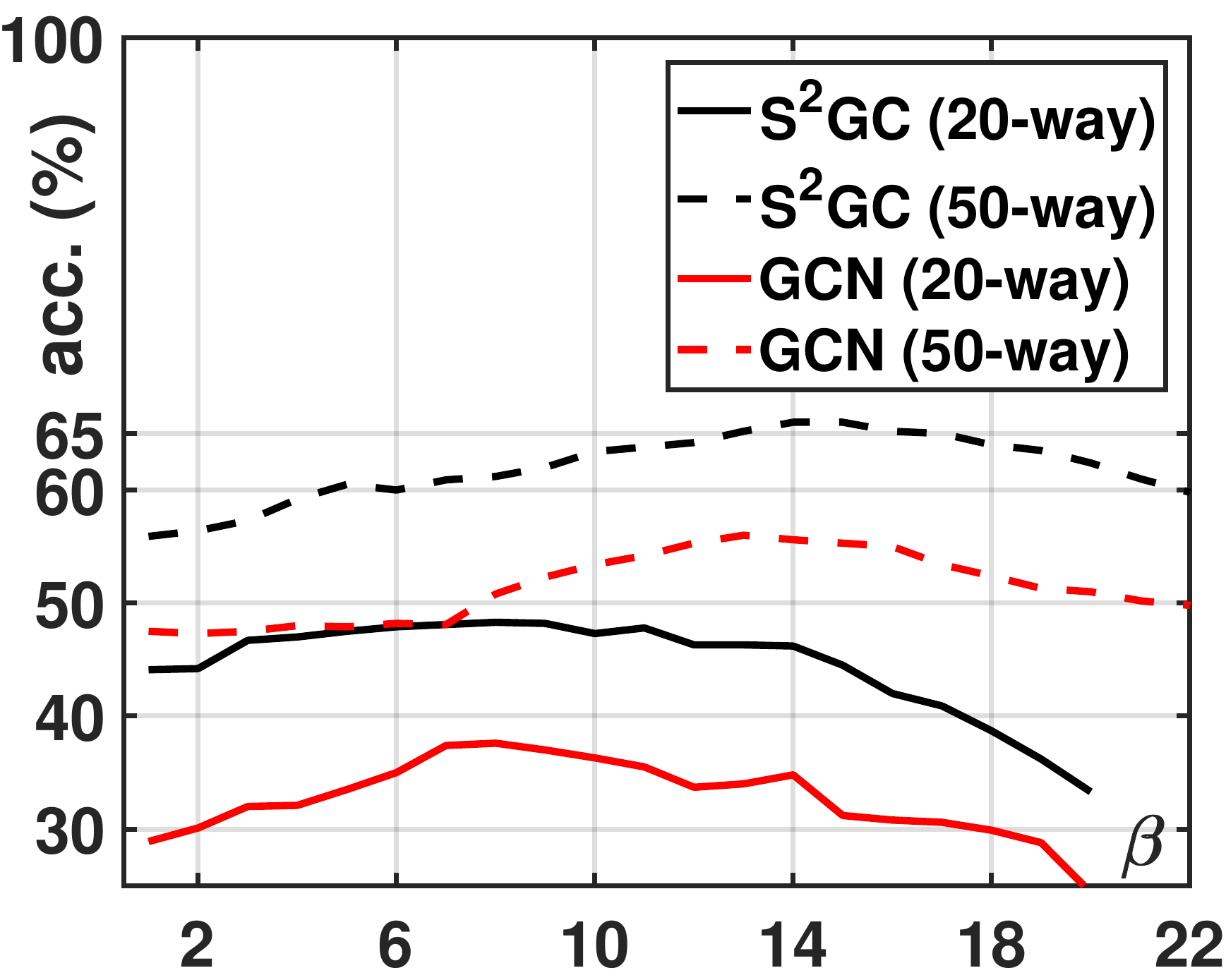}
\caption{\label{fig:beta} The impact of $\beta$ in loss function on NTU-60 with S$^2$GC and GCN.}
\end{minipage}
\end{figure}

\subsection{Ablation Study}

We start our experiments by investigating the GNN backbones (Appendix Sec.~\ref{backbone_selection}), camera viewpoint simulation and their hyper-parameters (Appendix Sec.~\ref{eva_alpha}, \ref{eva_layer}, \ref{stride-degree}). 

\definecolor{LightCyan}{rgb}{0.88,1,1}
\begin{table}[t]
\setlength{\tabcolsep}{0.25em}
\renewcommand{\arraystretch}{0.50}
\caption{Experimental results on NTU-60 (left) and NTU-120 (right) for different camera viewpoint simulations. Below the dashed line are ablated few variants of JEANIE.}
\begin{center}
\resizebox{0.95\linewidth}{!}{\begin{tabular}{ l c  c c c c  c  c  c c c c}
\toprule
& \multicolumn{5}{c}{NTU-60} & & \multicolumn{5}{c}{NTU-120}\\
\cline{2-6}
\cline{8-12}
\# Training Classes & 10 & 20 & 30 & 40 & 50 & & 20 & 40 & 60 & 80 & 100\\ 
\midrule
Euler simple ($K\!+\!K'$)&  54.3 & 56.2 & 60.4 &  64.0 & 68.1 & &  30.7 & 36.8 & 39.5 &  44.3 & 46.9\\ 
Euler ($K\!\times\!K'$)&  {\bf 60.8} & 67.4 & 67.5 &  70.3 & {\bf 75.0} & &  32.9 & 39.2 & 43.5 & 48.4 & 50.2 \\  
CamVPC ($K\!\times\!K'$)& 59.7 & {\bf 68.7} & {\bf 68.4} & {\bf 70.4} & 73.2 & & {\bf 33.1} & {\bf 40.8} & {\bf 43.7} & {\bf 48.4} & {\bf 51.4}\\
\hdashline
V(Euler)& 54.0 & 56.0 & 60.2 & 63.8 & 67.8 & & 30.6 & 36.7 & 39.2 & 44.0 & 47.0\\ 
2V(Euler simple)&  54.3 & 56.2 & 60.4 & 64.0 & 68.1& & 30.7  & 36.8 & 39.5 & 44.3 &  46.9\\  
2V(Euler)& 60.8 & 67.4 & 67.5 & 70.3 & 75.0& & 32.9 & 39.2 & 43.5 & 48.4 & 50.2 \\  
2V(CamVPC)& 59.7 & 68.7 & 68.4  & 70.4 & 73.2 & & 33.1 & 40.8 & 43.7 & 48.4 & 51.4 \\
\rowcolor{LightCyan}2V(CamVPC+crossval.)& 63.4 & 72.4 & 73.5  & 73.2 & 78.1 & & 37.2 & 43.0 & 49.2 & 50.0 & 55.2\\
\rowcolor{LightCyan}2V(CamVPC+crossval.)+Transf.& {\bf 65.0} & {\bf 75.2} & {\bf 76.7}  & {\bf 78.9} & {\bf 80.0} & & {\bf 38.5} & {\bf 44.1} & {\bf 50.3} & {\bf 51.2} & {\bf 57.0}\\
\bottomrule
\end{tabular}}
\label{ntu60_euler_camvpc}
\end{center}
\end{table}

\paragraph{Camera viewpoint simulations}. We choose 15 degrees as the step size for the viewpoints simulation. The ranges of camera azimuth/altitude are in [$-90^\circ$, $90^\circ$]. Where stated, we perform a grid search on camera azimuth/altitude with Hyperopt. 
Below, we explore the choice of the angle ranges for both horizontal and vertical views.  Fig.~\ref{fig:sgc_h_angles} and~\ref{fig:sgc_v_angles} (evaluations on the NTU-60 dataset) show that the angle range $[-45^\circ, 45^\circ]$ performs the best, and widening the range in both views does not increase the performance any further. 
Table~\ref{ntu60_euler_camvpc} (top) shows results for the chosen range $[-45^\circ,45^\circ]$ of camera viewpoint simulations.  
Euler simple ($K\!+\!K'$) denotes a simple concatenation of features from both horizontal and vertical views, whereas Euler/CamVPC($K\!\times\!K'$) represents the grid search of all possible views.
It shows that Euler angles for the viewpoint augmentation outperform Euler simple, and CamVPC (viewpoints of query sequences are generated by the stereo projection geometry)  outperforms Euler angles in almost all the experiments on NTU-60 and NTU-120. This proves the effectiveness of using the stereo projection geometry for the viewpoint augmentation. 
More baseline experiments with/without viewpoint alignment are in Appendix Sec. \ref{sup:vall}. 

\paragraph{Evaluation of $\beta$}. Figure~\ref{fig:beta} shows that if $\beta\!=\!8$ and 14, our loss function performs the best on 20- and 50-class protocol, respectively, on NTU-60 for the S$^2$GC and GCN backbone. Moreover, $\beta$ is not affected by backbone. 

\begin{table}[t]
\setlength{\tabcolsep}{0.25em}
\renewcommand{\arraystretch}{0.50}
\caption{Experimental results on NTU-60 (left) and NTU-120 (right) for $\iota$-max shift. $\iota$-max shift is the max. viewpoint shift from block to block in JEANIE.}
\begin{center}
\resizebox{0.65\linewidth}{!}{\begin{tabular}{ l  c  c  c  c  c  c c  c  c  c  c}
\toprule
& \multicolumn{5}{c}{NTU-60} & & \multicolumn{5}{c}{NTU-120}\\
\cline{2-6}
\cline{8-12}
 & 10 & 20 & 30 & 40 & 50 & & 20 & 40 & 60 & 80 & 100\\ 
\midrule
$\iota\!=\!1$& 60.8  & 70.7 & 72.5 & 72.9  & 75.2 & &  36.3 & 42.5 & 48.7 &  {\bf 50.0} & 54.8\\ 
\rowcolor{LightCyan}$\iota\!=\!2$& {\bf 63.8}  & {\bf 72.9} & {\bf 74.0} & {\bf73.4}  & {\bf 78.1} & & {\bf37.2}  & {\bf43.0} & {\bf 49.2} & {\bf50.0}  & {\bf 55.2}\\  
$\iota\!=\!3$&  55.2 & 58.9 & 65.7 & 67.1  & 72.5 & & 36.7  & {\bf 43.0} & 48.5 & 49.0  & 54.9\\ 
$\iota\!=\!4$&  54.5 & 57.8 & 63.5 &  65.2 & 70.4 & &  36.5 & 42.9 & 48.3 &  48.9 & 54.3\\ 
\bottomrule

\end{tabular}}
\label{ntu60_maxshift}
\end{center}
\end{table}

\paragraph{The $\iota$-max shift}. Table~\ref{ntu60_maxshift} shows the evaluations of $\iota$ for the maximum shift. We notice that $\iota\!=\!2$ yields the best results for all the experimental settings on both NTU-60 and NTU-120. Increasing $\iota$ does not help improve the performance. 

\begin{table}[t]
\begin{center}
\setlength{\tabcolsep}{0.25em}
\renewcommand{\arraystretch}{0.50}
\caption{
The impact of the number of frames $M$ in temporal block  under stride step $S$  on results  (NTU-60). $S\!=\!pM$, where $1\!-\!p$ describes the temporal block overlap percentage. Higher $p$ means fewer overlap frames between  temporal blocks.
}
\setlength{\tabcolsep}{2pt}
\resizebox{0.95\linewidth}{!}{
\renewcommand{\arraystretch}{0.60}
\begin{tabular}{ l c  c  c  c  c  c c  c c  c c c c c}
\toprule
 & \multicolumn{2}{c}{$S = M$} & & \multicolumn{2}{c}{$S = 0.8M$} && \multicolumn{2}{c}{$S = 0.6M$} && \multicolumn{2}{c}{$S = 0.4M$} && \multicolumn{2}{c}{$S = 0.2M$} \\
\cline{2-3}
\cline{5-6}
\cline{8-9}
\cline{11-12}
\cline{14-15}
 $M$ & 50-class & 20-class & & 50-class & 20-class && 50-class & 20-class && 50-class & 20-class && 50-class & 20-class \\
\midrule
5 & 69.0 & 55.7 && 71.8 & 57.2 && 69.2 & 59.6 && 73.0 & 60.8 && 71.2 & 61.2\\
6 & 69.4 & 54.0 && 65.4 & 54.1 && 67.8 & 58.0 && 72.0 & 57.8 && {\bf 73.0} & {\bf 63.0} \\
8 & 67.0 & 52.7 && 67.0 & 52.5 && {\bf 73.8}& {\bf 61.8} && 67.8 & 60.3 && 68.4 & 59.4 \\
10 & 62.2 & 44.5 && 63.6 & 50.9 && 65.2 & 48.4 && 62.4 & 57.0 && 70.4 & 56.7\\
15 & 62.0 & 43.5 && 62.6 & 48.9 && 64.7 & 47.9 && 62.4 & 57.2 && 68.3 & 56.7\\
30 & 55.6 & 42.8 && 57.2 & 44.8 && 59.2 & 43.9 && 58.8 & 55.3 && 60.2 & 53.8\\
45 & 50.0 & 39.8 && 50.5 & 40.6 && 52.3 & 39.9 && 53.0 & 42.1 && 54.0 & 45.2\\
\bottomrule
\end{tabular}}
\label{blockframe_overlap}
\end{center}
\end{table}

\paragraph{Block size and strides}. Table~\ref{blockframe_overlap} shows  evaluations of block size $M$ and stride $S$, and indicates %
that the best performance (both 50- and 20-class) is achieved  for smaller block size (frame count in the block) and smaller stride. %
Longer temporal blocks decrease the performance due to the temporal information not reaching the temporal alignment step. Our block encoder encodes each temporal block for learning the local temporal motions, and aggregate these block features finally to form the global temporal motion cues. Smaller stride helps capture more local motion patterns. %
Considering the computational cost and the performance, we choose $M\!=\!8$ and $S\!=\!0.6M$.

\paragraph{Euler \vs CamVPC}. Table~\ref{ntu60_euler_camvpc} (bottom) shows that using the viewpoint alignment simultaneously in two dimensions,  $x$ and $y$ for Euler angles, or azimuth and altitude the stereo projection geometry ({\em CamVPC}), improves the performance by 5-8\% compared to ({\em Euler simple}), a variant where the best viewpoint alignment path was chosen from the best alignment path along $x$ and the best alignment path along $y$. Euler simple is  better than  Euler with $y$ rotations only (({\em V}) includes rotations along $y$ while ({\em 2V}) includes rotations along two axes). 
Using  HyperOpt  \cite{bergstra2015hyperopt} to search for the best angle range in which we perform the viewpoint alignment ({\em CamVPC+crossval.}) improves results. Enabling the viewpoint alignment for support sequences yields extra improvement. With Transformer ({\em 2V+Transf.}), JEANIE boosts results by $\sim$ 2\%.

\begin{table}[t]
\setlength{\tabcolsep}{0.25em}
\renewcommand{\arraystretch}{0.50}
\caption{Results on NTU-60 (S$^2$GC backbone). Models use temporal alignment by soft-DTW or JEANIE (joint temporal-viewpoint alignment) except if indicated otherwise. 
}
\begin{center}
\resizebox{0.68\linewidth}{!}{\begin{tabular}{ l c c c c c c }
\toprule
\# Training Classes & & 10 & 20 & 30 & 40 & 50\\ 
\midrule

Each frame to frontal view & & 52.9 & 53.3 & 54.6 & 54.2 & 58.3\\
Each block  to frontal view & & 53.9 & 56.1 & 60.1 & 63.8 & 68.0\\

Traj. aligned baseline (video-level) & & 36.1 & 40.3 & 44.5 & 48.0 & 50.2\\
Traj. aligned baseline (block-level) & & 52.9 & 55.8 & 59.4 & 63.6 & 66.7\\
\hdashline

Matching Nets~\cite{f4Matching} & & 46.1 & 48.6 & 53.3 & 56.2 & 58.8\\
Matching Nets~\cite{f4Matching}+2V && 47.2 & 50.7 & 55.4 & 57.7 & 60.2\\
Prototypical Net~\cite{f1} && 47.2 & 51.1 & 54.3 & 58.9 & 63.0\\
Prototypical Net~\cite{f1}+2V && 49.8 & 53.1 & 56.7 & 60.9 & 64.3 \\
TAP~\cite{su2022temporal}
&& 54.2 & 57.3 & 61.7 & 64.7 & 68.3 \\
\hdashline
S$^2$GC (no soft-DTW) && 50.8 & 54.7 & 58.8 & 60.2 & 62.8\\ 
soft-DTW && 53.7 & 56.2 & 60.0 & 63.9 & 67.8 \\ 


(no soft-DTW)+Transf. & & 56.0 & 64.2 & 67.3 & 70.2 & 72.9\\
soft-DTW+Transf. && 57.3 & 66.1 & 68.8 & 72.3 & 74.0\\
\rowcolor{LightCyan}JEANIE+Transf.& & {\bf 65.0} & {\bf 75.2} & {\bf 76.7} & {\bf 78.9} & {\bf 80.0}\\
\bottomrule
\end{tabular}}
\label{ntu60results}
\end{center}
\vspace{-0.3cm}
\end{table}

\begin{table}[t]
\setlength{\tabcolsep}{0.25em}
\renewcommand{\arraystretch}{0.50}
\caption{Experimental results on NTU-120 (S$^2$GC backbone). Methods use temporal
alignment by soft-DTW or JEANIE (joint temporal-viewpoint
alignment) except VA \cite{Zhang_2017_ICCV,8630687} and other cited works. For VA$^*$, we used soft-DTW on temporal blocks while VA generated temporal blocks.}
\begin{center}
\resizebox{0.73\linewidth}{!}{\begin{tabular}{ l c c c c c c }
\toprule
\# Training Classes && 20  & 40  & 60  & 80  & 100 \\ 
\midrule
APSR~\cite{Liu_2019_NTURGBD120} && 29.1 & 34.8& 39.2 & 42.8 & 45.3 \\
SL-DML~\cite{2021dml}&& 36.7 & 42.4 & 49.0 & 46.4& 50.9 \\
Skeleton-DML~\cite{memmesheimer2021skeletondml} && 28.6 & 37.5 & 48.6 & 48.0 & 54.2 \\
\hdashline

Prototypical Net+VA-RNN(aug.)~\cite{Zhang_2017_ICCV} && 25.3 & 28.6 & 32.5 & 35.2 & 38.0 \\
Prototypical Net+VA-CNN(aug.)~\cite{8630687}&& 29.7 & 33.0 & 39.3 & 41.5 & 42.8 \\
Prototypical Net+VA-fusion(aug.)~\cite{8630687} && 29.8 & 33.2 & 39.5 & 41.7 & 43.0 \\
Prototypical Net+VA$^*$-fusion(aug.)~\cite{8630687} && 33.3 & 38.7 & 45.2 & 46.3 & 49.8 \\
TAP~\cite{su2022temporal}
& &31.2 & 37.7 & 40.9 & 44.5 &  47.3\\
\hdashline
S$^2$GC(no soft-DTW)&& 30.0 & 35.9 & 39.2 & 43.6 & 46.4 \\ 
soft-DTW && 30.3 & 37.2 & 39.7 & 44.0 & 46.8 \\ 
(no soft-DTW)+Transf. && 31.2 & 37.5 & 42.3 & 47.0 & 50.1\\
soft-DTW+Transf. && 31.6 & 38.0 & 43.2 & 47.8 & 51.3 \\
{\it FVM}+Transf. && 34.5 & 41.9 & 44.2 & 48.7 & 52.0 \\
\rowcolor{LightCyan}JEANIE+Transf. && {\bf 38.5} & {\bf 44.1} & {\bf 50.3} & {\bf 51.2} & {\bf 57.0}\\
\bottomrule
\end{tabular}}
\label{ntu120results}
\end{center}
\end{table}

\subsection{Comparisons With the State-of-the-Art Methods}
\noindent{\bf One-shot action recognition (NTU-60)}. 
Table~\ref{ntu60results} shows that aligning query and support trajectories by the angle of torso 3D joint, denoted  ({\em Traj. aligned baseline}) is not very powerful, as alluded to in Figure \ref{fig:sq-match} (top). Aligning piece-wise parts (blocks) is better than aligning entire trajectories. In fact, aligning individual frames by torso to the frontal view ({\em Each frame to frontal view}) and aligning block average of torso direction to the frontal view ({\em Each block  to frontal view})) were marginally better. We note these baselines use soft-DTW. We show more comparisons in Appendix Sec. \ref{sup:morebase}. Our JEANIE with Transformer ({\em JEANIE+Transf.}) outperforms soft-DTW with Transformer ({\em soft-DTW+Transf.}) by 7.46\% on average.

\paragraph{One-shot action recognition (NTU-120)}.  Table~\ref{ntu120results} shows that JEANIE  outperforms recent SL-DML and Skeleton-DML by 6.1\% and 2.8\% respectively (100 training classes). 
For comparisons, we extended the view adaptive neural networks~\cite{8630687} by combining them with Prototypical Net~\cite{f1}.
VA-RNN+VA-CNN~\cite{8630687} uses 0.47M+24M parameters with random rotation augmentations while JEANIE uses 0.25--0.5M params. Their {\em rotation}+{\em translation} keys are not proven to perform smooth optimal alignment as JEANIE. In contrast, $d_\text{JEANIE}$ performs jointly a smooth viewpoint-temporal alignment via a principled transportation plan ($\geq$3 dim. space) by design. Their use Euler angles which are a worse option than the camera projection of JEANIE. We notice that ProtoNet+VA backbones is 12\% worse than our JEANIE. Even if we split skeletons into blocks to let soft-DTW perform temporal alignment of prototypes and query, 
JEANIE is still 4--6\% better. \lei{JEANIE outperforms FVM by 2-4\%. This shows that seeking jointly the best temporal-viewpoint alignment is more valuable than considering viewpoint alignment as a local task (free range alignment per each step of soft-DTW).}

\begin{table}[t]
	\centering
	\setlength{\tabcolsep}{0.25em}
\renewcommand{\arraystretch}{0.50}
	\caption{Experiments on 2D and 3D Kinetics-skeleton. Note that we have no results on JEANIE or FVM
	~for 2D coordinates (aligning viewpoints is an operation in 3D).}
	\label{kinetics_results}  
	\resizebox{0.60\linewidth}{!}{\begin{tabular}{cccccc}  
		\toprule
		&S$^2$GC & \multirow{2}{*}{soft-DTW} & \multirow{2}{*}{\it FVM} & \multirow{2}{*}{JEANIE} & JEANIE\\
		&(no soft-DTW) &  &  &  & +Transf.\\%
		\midrule
		2D skel. & 32.8 & 34.7 & - & - & -\\
		\rowcolor{LightCyan}3D skel. & 35.9 & 39.6 & 44.1 & {\bf 50.3} & {\bf 52.5}\\
		\bottomrule
	\end{tabular}}
\end{table}

\paragraph{JEANIE on the Kinetics-skeleton}. We evaluate our proposed model on both 2D and 3D Kinetics-skeleton. We split the whole dataset into 200 actions for training, and the rest half for testing. As we are unable to estimate the camera location, we simply use Euler angles for the camera viewpoint simulation. 
Table~\ref{kinetics_results} shows that using 3D skeletons outperforms the use of 2D skeletons by 3-4\%, and  JEANIE outperforms the baseline (temporal alignment only) and Free Viewpoint Matching (FVM, for every step of DTW, seeks the best local viewpoint alignment, thus realizing non-smooth temporal-viewpoint path in contrast to JEANIE) by around 5\% and 6\%,  respectively. With the transformer, JEANIE  further boosts results by  2\%.

\begin{table}[t]
\begin{center}
\setlength{\tabcolsep}{0.25em}
\renewcommand{\arraystretch}{0.50}
\caption{Experiments  on the UWA3D Multiview Activity II.
}
\resizebox{0.92\linewidth}{!}{\begin{tabular}{ l  cccccc c  c  c  c  c c  c  c  c  c c  c  c  }
\toprule
\!\!\!Training view\!\!\! & & \multicolumn{2}{c}{$V_1$ \& $V_2$} & & \multicolumn{2}{c}{$V_1$ \& $V_3$} & & \multicolumn{2}{c}{$V_1$ \& $V_4$} &&  \multicolumn{2}{c}{$V_2$ \& $V_3$} & & \multicolumn{2}{c}{$V_2$ \& $V_4$} &&  \multicolumn{2}{c}{$V_3$ \& $V_4$} & \multirow{2}{*}{Mean}\\
\cline{1-1}
\cline{3-4}
\cline{6-7}
\cline{9-10}
\cline{12-13}
\cline{15-16}
\cline{18-19}
\!\!\!Testing view & & $V_3$ & $V_4$ & & $V_2$ & $V_4$ && $V_2$ & $V_3$ && $V_1$ & $V_4$ && $V_1$ & $V_3$ && $V_1$ & $V_2$ & {}\\
\midrule
GCN & &36.4 & 26.2 &&20.6 & 30.2 && 33.7 & 22.4 && 43.1 & 26.6 && 16.9 & 12.8 && 26.3 & 36.5 & 27.6 \\
\hdashline
SGC&&40.9&60.3&&44.1&52.6&&48.5&38.7&&50.6&52.8&&52.8&37.2&&57.8&49.6&48.8\\
+soft-DTW &&43.9&60.8&&48.1&54.6&&52.6&45.7&&54.0&58.2&&56.7&40.2&&60.2&51.1&52.2\\
\rowcolor{LightCyan}+JEANIE &&47.0&62.8&&50.4&57.8&&53.6&47.0&&57.9&62.3&&57.0&44.8&&61.7&52.3&54.6\\
\hdashline
APPNP&&42.9&61.9&&47.8&58.7&&53.8&44.0&&52.3&60.3&&55.1&38.2&&58.3&47.9&51.8\\
+soft-DTW &&44.3&63.2&&50.7&62.3&&53.9&45.0&&56.9&62.8&&56.4&39.3&&60.1&51.9&53.9\\
\rowcolor{LightCyan}+JEANIE &&46.8&64.6&&51.3&65.1&&54.7&46.4&&58.2&65.1&&58.8&43.9&&60.3&52.5&55.6\\
\hdashline
S$^2$GC&&45.5&64.4&&46.8&61.6&&49.5&43.2&&57.3&61.2&&51.0&42.9&&57.0&49.2&52.5\\
+soft-DTW &&48.2&67.2&&51.2&67.0&&53.2&46.8&&62.4&66.2&&57.8&45.0&&62.2&53.0&56.7\\
+FVM &&50.7&68.8&&56.3&69.2&&55.8&47.1&&63.7&68.8&&62.5&51.4&&63.8&55.7&59.5\\
\rowcolor{LightCyan}+JEANIE &&{\bf 55.3}&{\bf 70.2}&&{\bf 61.4}&{\bf 72.5}&&{\bf 60.9}&{\bf 50.8}&&{\bf 66.4}&{\bf 73.9}&&{\bf 68.8}&{\bf 57.2}&&{\bf 66.7}&{\bf 60.2}&{\bf 63.7}\\
\bottomrule
\end{tabular}}
\label{uwa3dmresults}
\end{center}
\end{table}
\begin{table}[t]
\setlength{\tabcolsep}{0.25em}
\renewcommand{\arraystretch}{0.50}
\caption{Results on NTU-120 (multiview classification). 
Baseline is soft-DTW + S$^2$GC. 
}
\begin{center}
\resizebox{0.76\linewidth}{!}{\begin{tabular}{ l c c c c c c c c }
\toprule
Training view & & bott. & bott. & bott.\& cent. & & left & left & left \& cent.\\
\cline{1-1}
\cline{3-5}
\cline{7-9}
Testing view & & cent. & top & top && cent. & right & right \\
\midrule
100/same 100 (baseline)& & 74.2 & 73.8 & 75.0 && 58.3 & 57.2 & 68.9\\
100/same 100 ({\it FVM})&& 79.9 & 78.2 & 80.0 && 65.9 & 63.9 & 75.0\\

\rowcolor{LightCyan}100/same 100 ({\it JEANIE})&& {\bf 81.5} & {\bf 79.2} & {\bf 83.9} && {\bf 67.7} & {\bf 66.9} & {\bf 79.2}\\
\hdashline
100/novel 20 (baseline) && 58.2 & 58.2 & 61.3 && 51.3 & 47.2 & 53.7 \\
100/novel 20 ({\it FVM}) && 66.0 & 65.3 & 68.2 && 58.8 & 53.9 & 60.1 \\
\rowcolor{LightCyan}100/novel 20 ({\it JEANIE}) && {\bf 67.8} & {\bf 65.8} & {\bf 70.8} & &{\bf 59.5} & {\bf 55.0} & {\bf 62.7} \\
\bottomrule
\end{tabular}}
\label{ntu120results_view}
\end{center}
\end{table}

\paragraph{Few-shot multiview classification}. Table~\ref{uwa3dmresults} (UWA3D Multiview Activity II) shows that adding temporal alignment to SGC, APPNP and S$^2$GC improves the performance, and the big performance gain is obtained by further adding the viewpoint alignment. As this dataset is challenging in recognizing the actions from a novel view point, our proposed method performs consistently well on all different combinations of training/testing viewpoint variants. This is predictable as our method aligns both temporal and camera viewpoints which allows a robust classification. \lei{JEANIE outperforms  FVM by 4.2\%, and outperforms the baseline (with temporal alignment only) by 7\% on average.} 

Table~\ref{ntu120results_view} (NTU-120) shows that adding more camera viewpoints to the training process helps the multiview classification. Using bottom and center views for training and top view for testing, or using left and center views for training and the right view for testing yields 4\% gain (`{\em same 100}' means the same train/test classes but different views). Testing on 20 novel classes (`{\em novel 20}'  never used in training) yields 62.7\% and 70.8\% for multiview classification in horizontal and vertical camera viewpoints, respectively.

\section{Conclusions}

We have proposed a Few-shot Action Recognition (FSAR) approach for learning on 3D skeletons via JEANIE. We have demonstrated that the joint alignment of temporal blocks and simulated viewpoints of skeletons between support-query sequences is efficient in the meta-learning setting where the alignment has to be performed on new action classes under the low number of samples. Our experiments have shown that using the stereo camera geometry is more efficient than simply generating multiple views by Euler angles in the meta-learning regime. Most importantly, we have introduced a novel FSAR approach that learns on articulated 3D body joints.

\section*{Acknowledgements}
We thank Dr. Jun Liu (SUTD) for discussions on FSAR for 3D skeletons, and CSIRO’s Machine Learning and
Artificial Intelligence Future Science Platform (MLAI FSP).



\bibliographystyle{splncs04}
\bibliography{egbib}

\appendix

\pagestyle{headings}

\title{Temporal-Viewpoint Transportation Plan for Skeletal Few-shot Action Recognition \\--- Supplementary Material ---}

\author{Lei Wang\inst{\dagger, \S}\orcidlink{0000-0002-8600-7099} \and
Piotr Koniusz\inst{\S,\dagger}\orcidlink{0000-0002-6340-5289}}
\authorrunning{Wang and Koniusz}
\titlerunning{Temporal-Viewpoint Transportation Plan for Skeletal FSAR (Appendix)}
%
\institute{$^{\dagger}$Australian National University \;
   $^\S$Data61/CSIRO\\
   $^\S$firstname.lastname@data61.csiro.au
}

\maketitle

\thispagestyle{empty}

\setcounter{table}{8}
\setcounter{equation}{5}
\setcounter{figure}{6}

\definecolor{tealgreen}{rgb}{0.0, 0.51, 0.5}
\renewcommand{\lei}{\textcolor{tealgreen}}


Below are additional derivations, evaluations and illustrations of our method.

\section{Prerequisites}
\label{supp:prereq}

\paragraph{Euler angles} \citelatex{eulera_supp} are defined as successive planar rotation angles around $x$, $y$, and $z$ axes.  For 3D coordinates, we have  the following rotation matrices ${\bf R}_x$, ${\bf R}_y$ and ${\bf R}_z$:
%
%
\begin{align}
\left[\arraycolsep=1.4pt\def\arraystretch{0.5}\begin{array}{ccc}
1 & 0 & 0\\
0 & \text{cos}\theta_x & \text{sin}\theta_x\\
0 & -\text{sin}\theta_x & \text{cos}\theta_x
\end{array} 
\right ],
\left[\arraycolsep=1.4pt\def\arraystretch{0.5}\begin{array}{ccc}
\text{cos}\theta_y & 0 & -\text{sin}\theta_y\\
0 & 1 & 0\\
\text{sin}\theta_y & 0 & \text{cos}\theta_y
\end{array} 
\right],
\left[\arraycolsep=1.4pt\def\arraystretch{0.5} \begin{array}{ccc}
\text{cos}\theta_z & \text{sin}\theta_z &  0\\
-\text{sin}\theta_z & \text{cos}\theta_z & 0\\
0 & 0 & 1
\end{array} 
\right]
\end{align}

\noindent As the resulting composite rotation matrix  depends on the order of  rotation axes, \ie, $\mathbf{R}_x\mathbf{R}_y\mathbf{R}_z\!\neq\!\mathbf{R}_z\mathbf{R}_y\mathbf{R}_x$, we also investigate    the algebra of stereo projection.

\paragraph{Stereo projections} \citelatex{sterproj_supp}. 
Suppose we have a rotation matrix ${\bf R}$ and a translation vector ${\bf t}\!=\![t_x, t_y, t_z]^T$ between left/right cameras (imagine some non-existent stereo camera). Let ${\bf M}_l$ and ${\bf M}_r$ be  the intrinsic matrices of the left/right cameras.  
Let ${\bf p}_l$ and ${\bf p}_r$ be coordinates of the left/right camera. As the origin of the right camera in the left camera coordinates is ${\bf t}$,  we have: ${\bf p}_r\!=\!{\bf R}({\bf p}_l\!-\!{\bf t})$ and  
$({\bf p}_l\!-\!{\bf t})^T\!=\!({\bf R}^{T}{\bf p}_r)^T$. 
The plane  (polar surface) formed by all points passing through ${\bf t}$ can be expressed by $({\bf p}_l\!-\!{\bf t})^T({\bf p}_l\!\times\!{\bf t})\!=\!0$. Then, ${\bf p}_l\!\times \!{\bf t}\!=\!{\bf S}{\bf p}_l$ where ${\bf S}\!=\! \left[\arraycolsep=1.4pt\def\arraystretch{0.5}\begin{array}{ccc}
0 & -t_z & t_y \\
t_z & 0 & -t_x\\
-t_y & t_x & 0
\end{array} 
\right]$. 
Based on the above equations, we obtain ${{\bf p}_r}^T{\bf R}{\bf S}{\bf p}_l\!=\!0$, and note that ${\bf R}{\bf S}\!=\!{\bf E}$ is the Essential Matrix, and ${{\bf p}^T_r}{\bf E} {\bf p}_l\!=\!0$ describes the relationship for the same physical point under the left and right camera coordinate system. 
As ${\bf E}$ has no internal inf. about the camera, and ${\bf E}$ is based on the camera coordinates, we  use a fundamental matrix {\bf F} that describes the relationship for the same physical point under the camera pixel coordinate system. 
The relationship between the pixel  and camera coordinates is: ${\bf p}^*\!=\!{\bf M}{\bf p}'$ and ${{\bf p}'_r}^T{\bf E} {\bf p}'_l\!=\!0$.

Now, suppose the pixel coordinates of ${\bf p}'_l$ and ${\bf p}'_r$ in the pixel coordinate system are ${\bf p}^*_{l}$ and ${\bf p}^*_{r}$, then we can write ${{\bf p}^*_{r}}^T({\bf M}_r^{-1})^T{\bf E}{\bf M}_l^{-1}{\bf p}^*_{l}\!=\!0$, where  ${\bf F}\!=\!({\bf M}_r^{-1})^T{\bf E}{\bf M}_l^{-1}$ is the fundamental matrix. Thus, the relationship for the same point in the pixel coordinate system of the left/right camera is:
%
\begin{equation}
{{\bf p}^*_{r}}^{T}{\bf F}{\bf p}^*_{l}\!=\!0.
\label{eq:f_matrix}
\end{equation}

\noindent We treat 3D body joint coordinates as ${\bf p}^*_{l}$. Given ${\bf F}$ (estimation of ${\bf F}$ is  explained in {\bf stereo projections} of Sec.~\ref{sec:exper} in the main paper), we obtain their coordinates ${\bf p}^*_{r}$ in the new view.

\paragraph{GNN notations.} 
Firstly, let $G\!=\!(\bf{V}, {\bf E})$ be a graph with the vertex set $\bf{V}$  with nodes $\{v_1, \cdots, v_n\}$, and ${\bf E}$ are edges of the graph. Let ${\bf A}$ and ${\bf D}$ be the adjacency and diagonal degree matrix, respectively. Let $\tilde{\bf A}\!=\!{\bf A}\!+\!{\bf I}$ be the adjacency matrix with self-loops (identity matrix) with the corresponding diagonal degree matrix $\tilde{\bf D}$ such that $\tilde{D}_{ii}\!=\!\sum_j ({\bf A}^{ij}\!+\! {\bf I}^{ij})$. 
Let ${\bf S}\!=\!\tilde{\bf D}^{-\frac{1}{2}} \tilde{\bf A}\tilde{\bf D}^{-\frac{1}{2}}$ be the normalized adjacency matrix with added self-loops. For the $l$-th layer, we use ${\bf \Theta}^{(l)}$ to denote the learnt weight matrix, and ${\bf \Phi}$ to denote the outputs from the graph networks. Below, we list backbones used by us.

\paragraph{GCN}~\citelatex{kipf2017semi_supp}. GCNs learn the feature representations for the features $\mathbf{x}_i$ of each node over multiple layers. For the $l$-th layer, we denote the input  by  ${\bf H}^{(l-1)}$ and the output by ${\bf H}^{(l)}$. Let the input (initial) node representations be ${\bf H}^{(0)}\!=\! {\bf X}$. 
For an $L$-layer GCN, the output representations are given by:
\begin{equation}
    {\bf \Phi_\text{GCN}}\!=\!{\bf S}{\bf H}^{(L-1)}{\bf \Theta}^{(L)} \text{ where } {\bf H}^{(l)}\!\!=\!\text{ReLU}({\bf S}{\bf H}^{(l-1)}{\bf \Theta}^{(l)}).
\end{equation}

\paragraph{APPNP}~\citelatex{johannes2019iclr_supp}. The Personalized Propagation of Neural Predictions (PPNP) and its fast approximation, APPNP, are based on the  personalized PageRank. Let  ${\bf H}^{(0)}\!=\!f_{\bf \Theta}(X)$ be the input to APPNP, where $f_{\bf \Theta}$ can be an MLP with parameters  ${\bf \Theta}$. Let the output of the $l$-th layer be ${\bf H}^{(l)}\!=\! (1-\alpha){\bf S}{\bf H}^{(l-1)}\!+\!\alpha{\bf H}^{(0)}$, where $\alpha$ is the teleport (or restart) probability  in range $(0, 1]$. For an $L$-layer APPNP, we have:
\begin{equation}
    {\bf \Phi_\text{APPNP}}\!=\!{(1\!-\!\alpha)\bf S}{\bf H}^L\!+\!\alpha{\bf H^{(0)}}.
\end{equation}

\paragraph{SGC}~\citelatex{felix2019icml_supp} { \&} {{\bf S$^2$GC}}~\citelatex{hao2021iclr_supp}. 
SGC captures the  $L$-hops neighborhood in the graph by the $L$-th power of the transition matrix used as a spectral filter. For an $L$-layer SGC,  
we obtain:
\begin{equation}
    {\bf \Phi_\text{SGC}}\!=\!{\bf S}^L{\bf X}{\bf \Theta}.
\end{equation}

Based on a modified Markov Diffusion Kernel, Simple Spectral Graph Convolution (S$^2$GC) is the summation over $l$-hops, $l\!=\!1,\cdots,L$.  The output of S$^2$GC is: 
\begin{equation}
    {\bf \Phi_\text{S$^2$GC}} \!=\!  \frac{1}{L}\sum_{l=1}^{L}((1\!-\!\alpha){\bf S}^l{\bf X}\!+\!\alpha{\bf X}){\bf \Theta}.
\end{equation}

\paragraph{Soft-DTW} \citelatex{marco2011icml_supp,marco2017icml_supp}. Dynamic Time Warping can be seen as a specialized  case of the Wasserstein metric, under specific transportation plan. Soft-DTW is defined as:
\begin{align}
& d_{\text{DTW}}(\mPsi,\mPsi')\!=\!\softming\limits_{\mA\in\tAnb_{\tau,\tau'}}\left\langle\mA,\mD(\mPsi,\mPsi')\right\rangle,\\
& \text{ where } \softming(\valpha)\!=\!-\gamma\!\log\sum_i\exp(-\alpha_i/\gamma).
\end{align}
The binary  $\mA\!\in\!\tAnb_{\tau,\tau'}$ denotes a path within the transportation plan $\tAnb_{\tau,\tau'}$ which depends on lengths $\tau$ and $\tau'$ of sequences $\mPsi\!\equiv\![\vpsi_1,\cdots,\vpsi_\tau]\!\in\!\mbr{d'\times\tau}$, $\mPsi'\!\equiv\![{\vpsi'}_1,\cdots,{\vpsi'}_{\tau'}]\!\in\!\mbr{d'\times\tau'}$ and  $\mD\!\in\!\mbrp{\tau\times\tau'}\!\!\equiv\![d_{\text{base}}(\vpsi_m,\vpsi'_n)]_{(m,n)\in\idx{\tau}\times\idx{\tau'}}$, the matrix of distances, is evaluated for $\tau\!\times\!\tau'$ frame representations according to some base distance $d_{\text{base}}(\cdot,\cdot)$, \ie, the Euclidean or the RBF-induced distance. 
We make use of principles of soft-DTW. However, we design a joint alignment between temporal skeleton sequences and simulated skeleton viewpoints, an entirely novel proposal.

\section{Datasets and their statistics}
\label{app:ds}

Table \ref{datasets} contains statistics of datasets used in our experiments. 
Smaller datasets below are used for the backbone selection and ablations: 

\begin{itemize}

\item {\em{MSRAction3D}}~\citelatex{Li2010_supp} is an older AR datasets captured with the Kinect depth camera. It contains 20 human sport-related activities such as {\it jogging}, {\it golf swing} and {\it side boxing}.

\item {{\em 3D Action Pairs}}~\citelatex{Oreifej2013_supp} contains 6 selected pairs of actions that have very similar motion trajectories, \eg, {\it put on a hat} and {\it take off a hat}, {\it pick up a box} and {\it put down a box}, \etc. 

\item {{\em UWA3D Activity}}~\citelatex{RahmaniHOPC2014_supp} has 30 actions performed by 10 people of various height at different speeds in cluttered scenes. 
\end{itemize}

As MSRAction3D, 3D Action Pairs, and UWA3D Activity have not been used in FSAR, we created 10 training/testing splits, by choosing half of class concepts for training, and half for testing per split per dataset. Training splits were further subdivided for crossvalidation.

Sec.~\ref{small_proto} details the class concepts per split for small datasets.

\begin{table*}[t]
\caption{Seven publicly available benchmark datasets which we use for FSAR.}
\begin{center}
\resizebox{\textwidth}{!}{\begin{tabular}{ l  c c c c  c  c  c  c  c }
\toprule
 Datasets & & Year & Classes & Subjects & \#views & \#clips & Sensor & Modalities & \#joints \\ 
\midrule
MSRAction3D~& \citelatex{Li2010_supp} & 2010 & 20 & 10 & 1 & 567 & Kinect v1 & Depth + 3DJoints & 20\\
3D Action Pairs~& \citelatex{Oreifej2013_supp} & 2013 & 12 & 10 & 1 & 360 & Kinect v1 & RGB + Depth + 3DJoints & 20\\
UWA3D Activity~& \citelatex{RahmaniHOPC2014_supp} & 2014 & 30 & 10 & 1 & 701 & Kinect v1 & RGB + Depth + 3DJoints & 15\\
UWA3D Multiview Activity II& \citelatex{Rahmani2016_supp} & 2015 & 30 & 9 & 4 & 1,070 & Kinect v1 & RGB + Depth + 3DJoints & 15\\
NTU RGB+D~& \citelatex{Shahroudy_2016_NTURGBD_supp} & 2016 & 60 & 40 & 80 & 56,880 & Kinect v2 & RGB + Depth + IR + 3DJoints & 25\\
NTU RGB+D 120~& \citelatex{Liu_2019_NTURGBD120_supp} & 2019 & 120 & 106 & 155 & 114,480 & Kinect v2 & RGB + Depth + IR + 3DJoints & 25\\
Kinetics-skeleton~& \citelatex{stgcn2018aaai_supp} & 2018 & 400 & - & - & $\sim$ 300,000 & - & RGB + 2DJoints & 18 \\
\bottomrule
\end{tabular}}
\label{datasets}
\end{center}
\end{table*}

\section{Backbone selection and hyperparameter evaluation}
\label{backbone_hyperpara}

\begin{figure}[t]
\centering
\begin{subfigure}[b]{0.45\linewidth}
\includegraphics[trim=0 0 0 0, clip=true,width=0.99\linewidth]{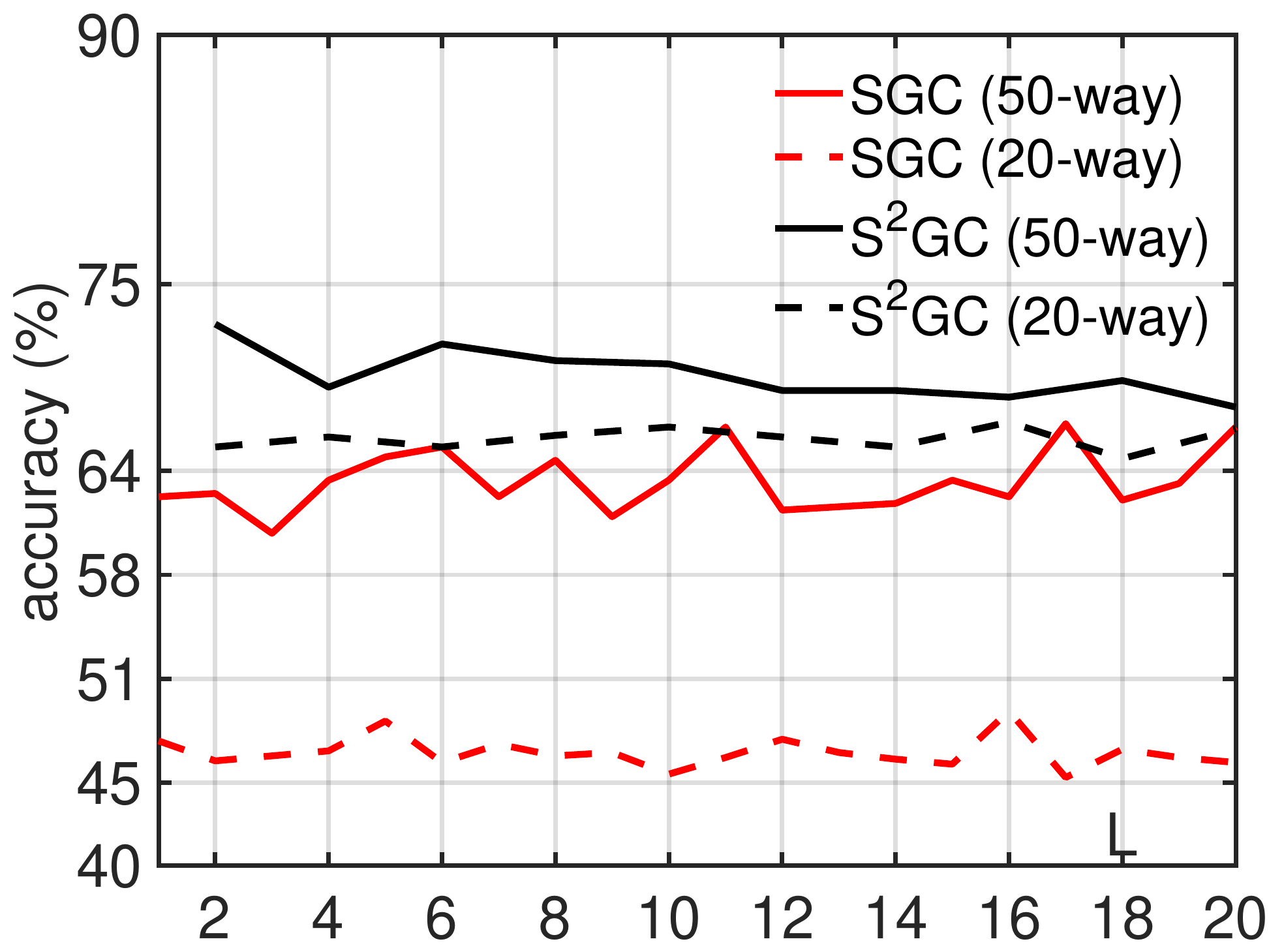}
\caption{\label{fig:degree}}
\end{subfigure}
\begin{subfigure}[b]{0.45\linewidth}
\includegraphics[trim=0 0 0 0, clip=true,width=0.99\linewidth]{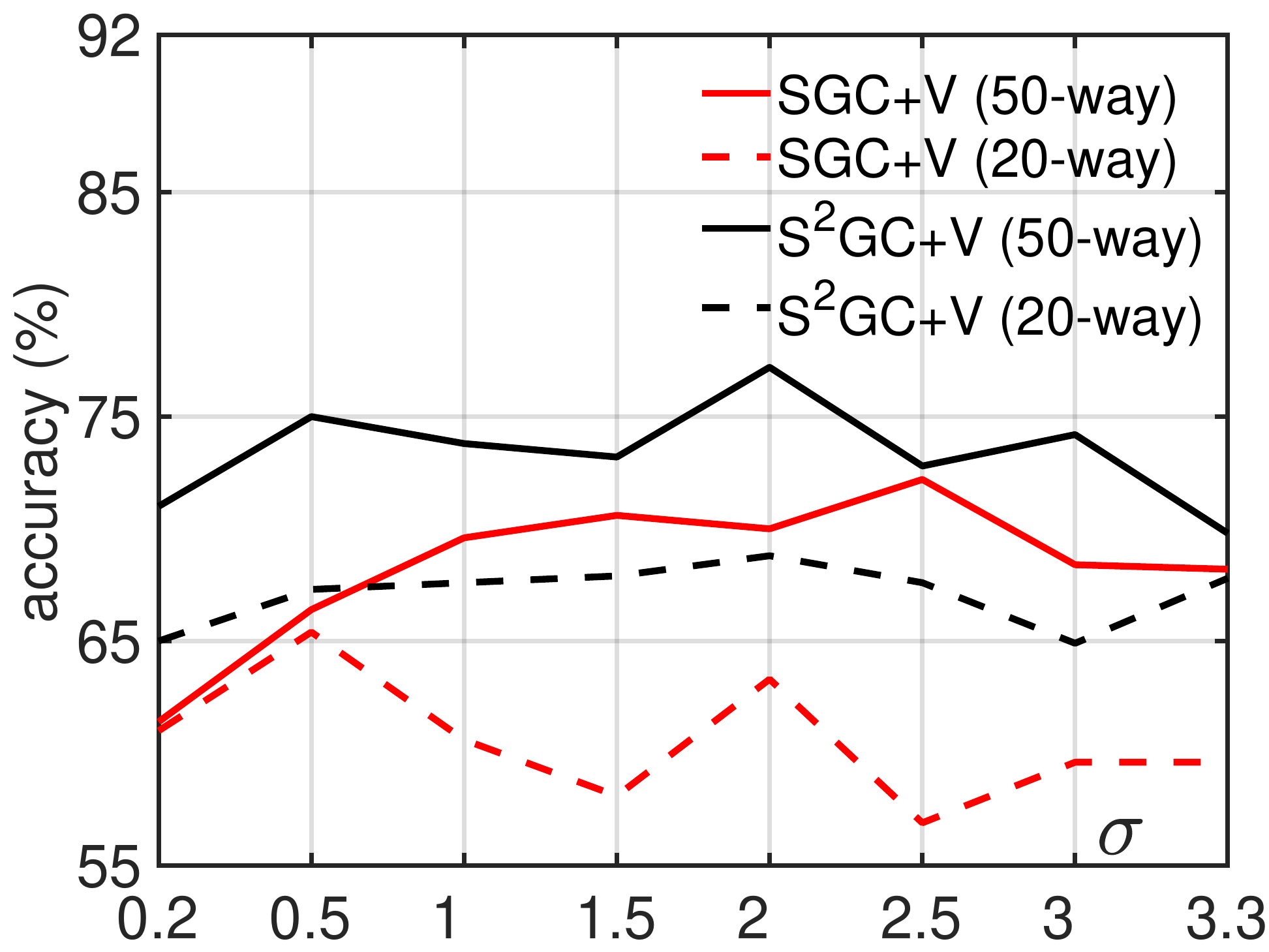}
\caption{\label{fig:sigma}}
\end{subfigure}
\caption{
Evaluations of $L$ and $\sigma$. (a): $L$ for SGC and S$^2$GC. (b): $\sigma$ of RBF distance for Eq.~\eqref{eq:d_jeanie} (SGC and S$^2$GC, NTU-60).
}
\label{fig:sigma_degree}
\end{figure}

\begin{figure}[t]
\centering
\begin{subfigure}[b]{0.45\linewidth}
\includegraphics[trim=0 0 0 0, clip=true,width=0.99\linewidth]{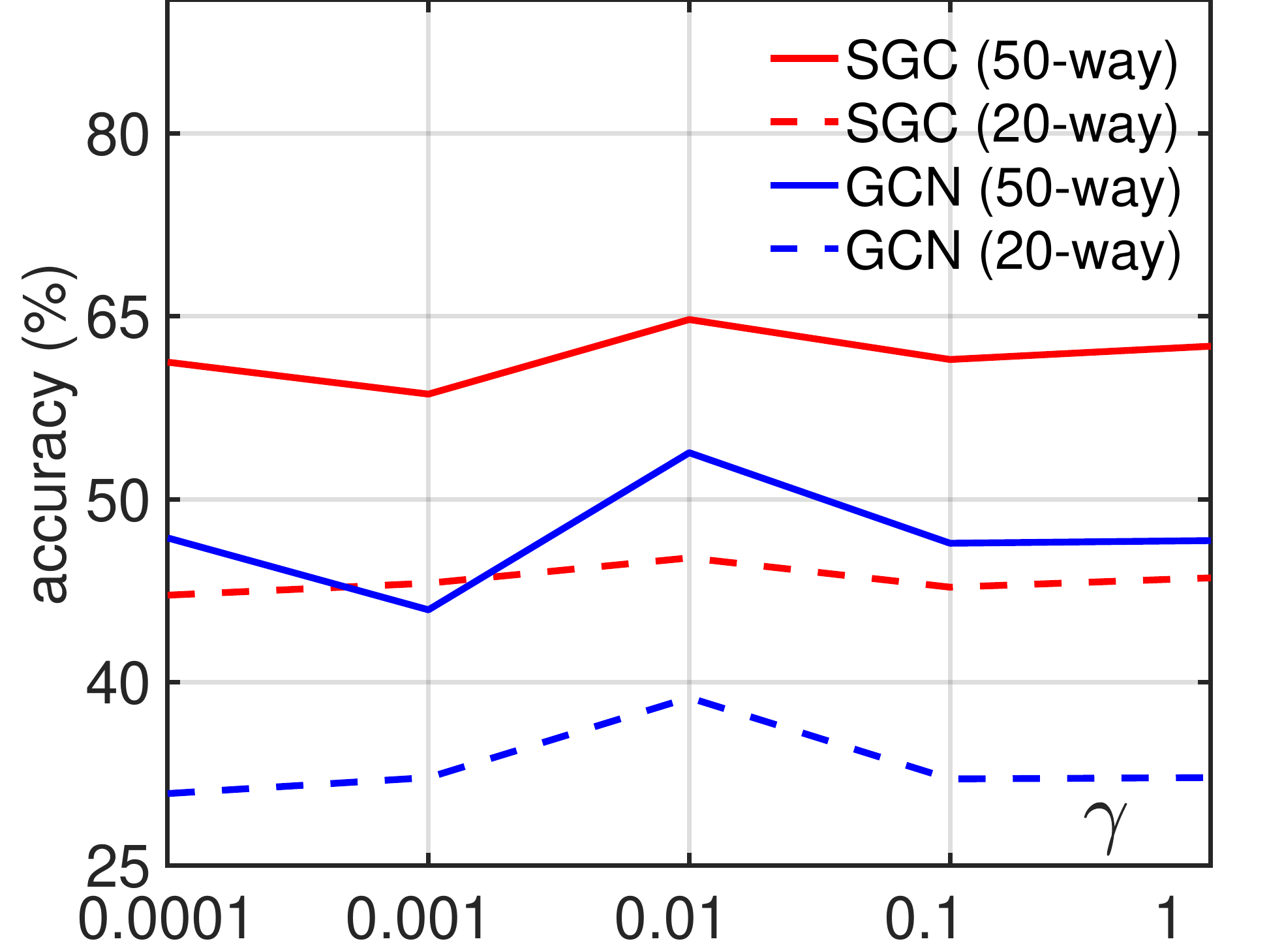}
\caption{\label{fig:sgc_gcn_gamma}}
\end{subfigure}
\begin{subfigure}[b]{0.45\linewidth}
\includegraphics[trim=0 0 0 0, clip=true,width=0.99\linewidth]{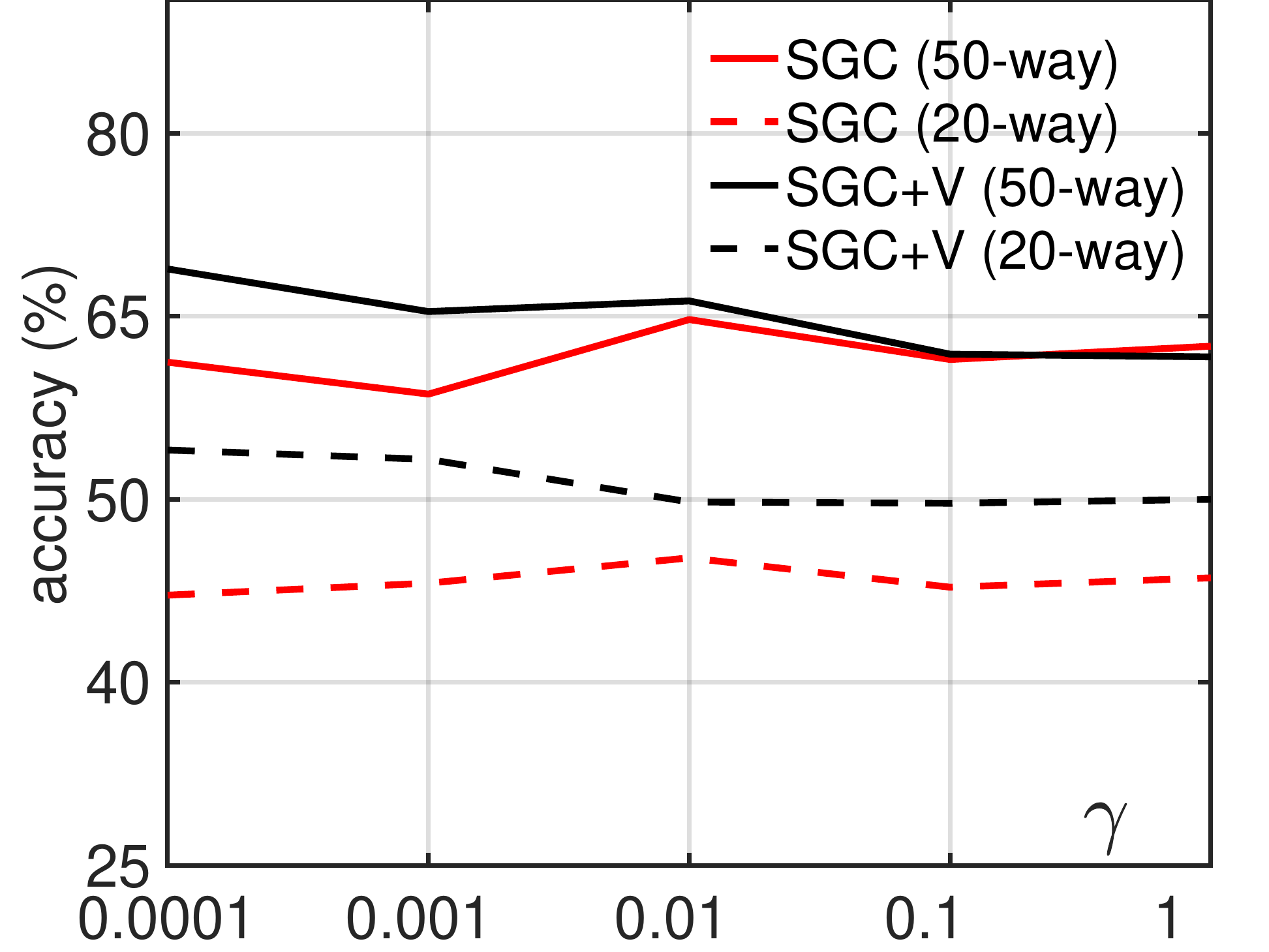}
\caption{\label{fig:sgc_vt_gamma}}
\end{subfigure}
\caption{
Evaluations \wrt $\gamma$. (a): $\gamma$ in Eq.~\eqref{eq:d_jeanie} with the temporal alignment alone.  (b): comparisons of temporal alignment alone \vs temporal-viewpoint alignment ({\em V}) on NTU-60.
}
\label{fig:gamma}
\end{figure}


\begin{table}[t]
\begin{center}
\caption{Evaluations of backbones on 5 datasets.}
\resizebox{\linewidth}{!}{
\renewcommand{\arraystretch}{0.9}
\setlength{\tabcolsep}{0.15em}
\begin{tabular}{cccccccccccc}
\toprule
\multirow{2}{*}{} & \multicolumn{2}{c}{MSRAction3D} & & 3DAct.Pairs & &  \multicolumn{2}{c}{UWA3DActivity} & & NTU-60 & & NTU-120\\
\cline{2-3}
\cline{5-5}
\cline{7-8}
\cline{10-10}
\cline{12-12}
 & 5-way & 10-way & & 5-way & & 5-way & 10-way & &  50-way & & 20-way \\
\midrule
GCN &  56.0 $\!\pm\!$ 1.3 &  37.6 $\!\pm\!$ 1.2 & &   - & & 55.4 $\!\pm\!$ 0.8  & 42.4 $\!\pm\!$ 0.8 & & 56.0 & &  -\\
SGC & 66.0 $\!\pm\!$ 1.1 &  48.3 $\!\pm\!$ 1.1 & &   69.0 $\!\pm\!$ 1.8 & & 56.4 $\!\pm\!$ 0.7 & 41.6 $\!\pm\!$ 0.6 & & 68.1 &&  30.7  \\
APPNP & 67.2 $\!\pm\!$ 0.8 & 58.1 $\!\pm\!$ 0.8 &  & 69.0 $\!\pm\!$ 2.0 & & 60.6 $\!\pm\!$ 1.5 & 42.4 $\!\pm\!$ 1.3 & & 68.5 & & 30.8 \\
S$^2$GC (Eucl.) & 68.8 $\!\pm\!$ 1.2  & 63.1 $\!\pm\!$ 0.9 & & 72.2 $\!\pm\!$ 1.8 & & 69.8 $\!\pm\!$ 0.7 & 58.3 $\!\pm\!$ 0.6 & & 75.6 & & 34.5  \\
\rowcolor{LightCyan}S$^2$GC (RBF) & \textbf{73.2 $\!\pm\!$ 0.9} & \textbf{64.6 $\!\pm\!$ 0.8} & &  \textbf{75.6 $\!\pm\!$ 2.1} & & \textbf{76.4 $\!\pm\!$ 0.7} & \textbf{58.9 $\!\pm\!$ 0.7} & & \textbf{78.1} &&   \textbf{36.2} \\
\bottomrule
\end{tabular}}
\label{backboneresults}
\end{center}
\end{table}

\subsection{Backbone selection}
\label{backbone_selection}

We conduct experiments on 4 GNN backbones listed in Table~\ref{backboneresults}. S$^2$GC performs the best on all  datasets including large-scale NTU-60 and NTU-120, APPNP outperforms  SGC, and SGC outperforms GCN. We note that using the RBF-induced distance for $d_{base}(\cdot,\cdot)$ of DTW outperforms the Euclidean distance. Fig.~\ref{fig:sigma_degree} shows a comparison of using SGC and S$^2$GC on NTU-60. As shown in the figure that using S$^2$GC performs better than SGC for both 50-way and 20-way settings. We also notice that results \wrt the number of layers $L$  are more  stable for S$^2$GC than SGC. We choose $L\!=\!6$ for our experiments. Fig.~\ref{fig:sigma} shows  evaluations of $\sigma$ for the RBF-induced distance for both SGC and S$^2$GC. As $\sigma\!=\!2$ in S$^2$GC achieves the highest performance (both 50-way and 20-way), we choose S$^2$GC as the backbone  and $\sigma\!=\!2$ for the experiments.

\begin{figure}[t]
\centering
\begin{subfigure}[b]{0.45\linewidth}
\includegraphics[trim=0 0 0 0, clip=true,width=0.99\linewidth]{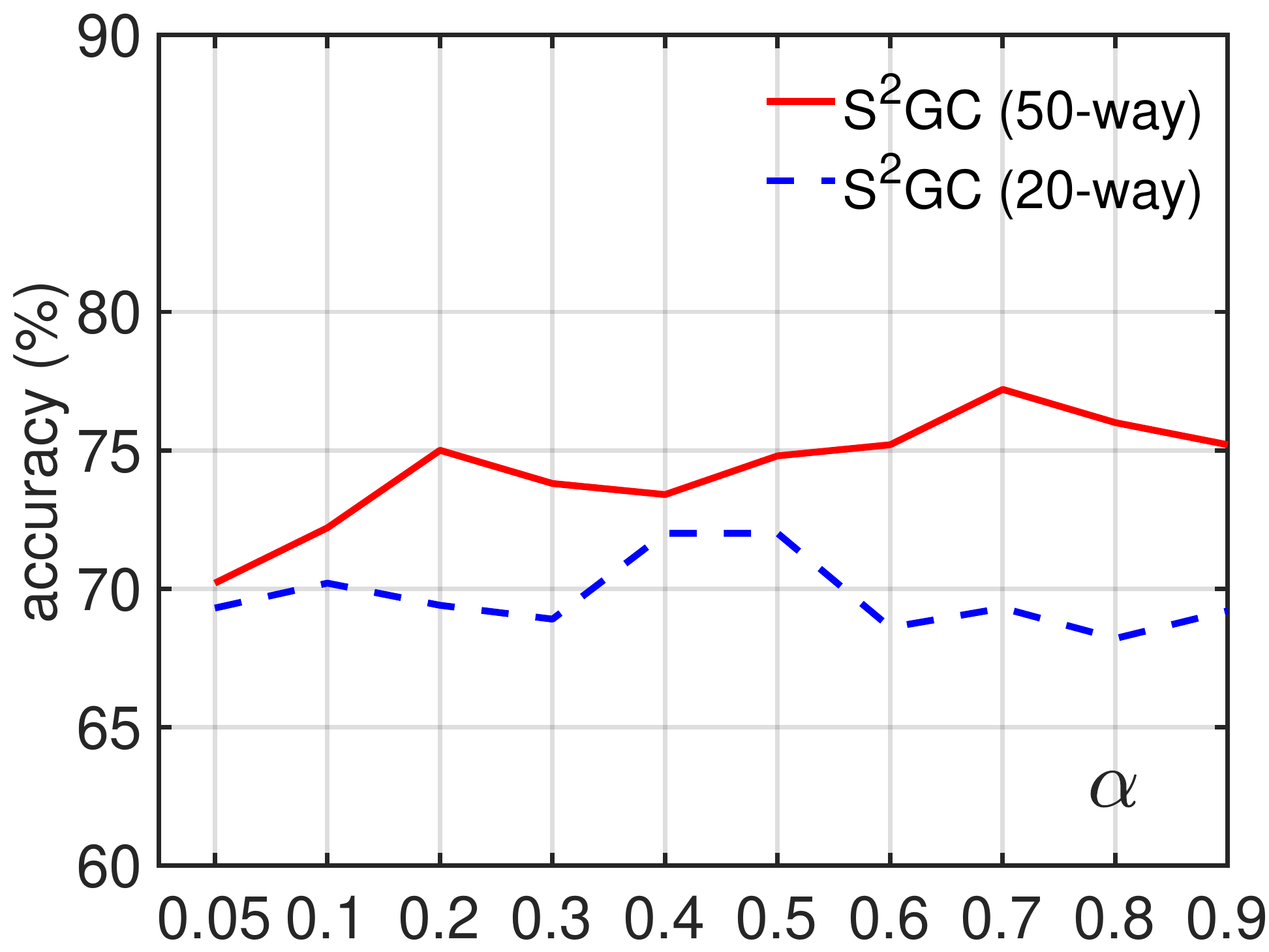}
\caption{\label{fig:ssgc_alpha}}
\end{subfigure}
\begin{subfigure}[b]{0.45\linewidth}
\includegraphics[trim=0 0 0 0, clip=true,width=0.99\linewidth]{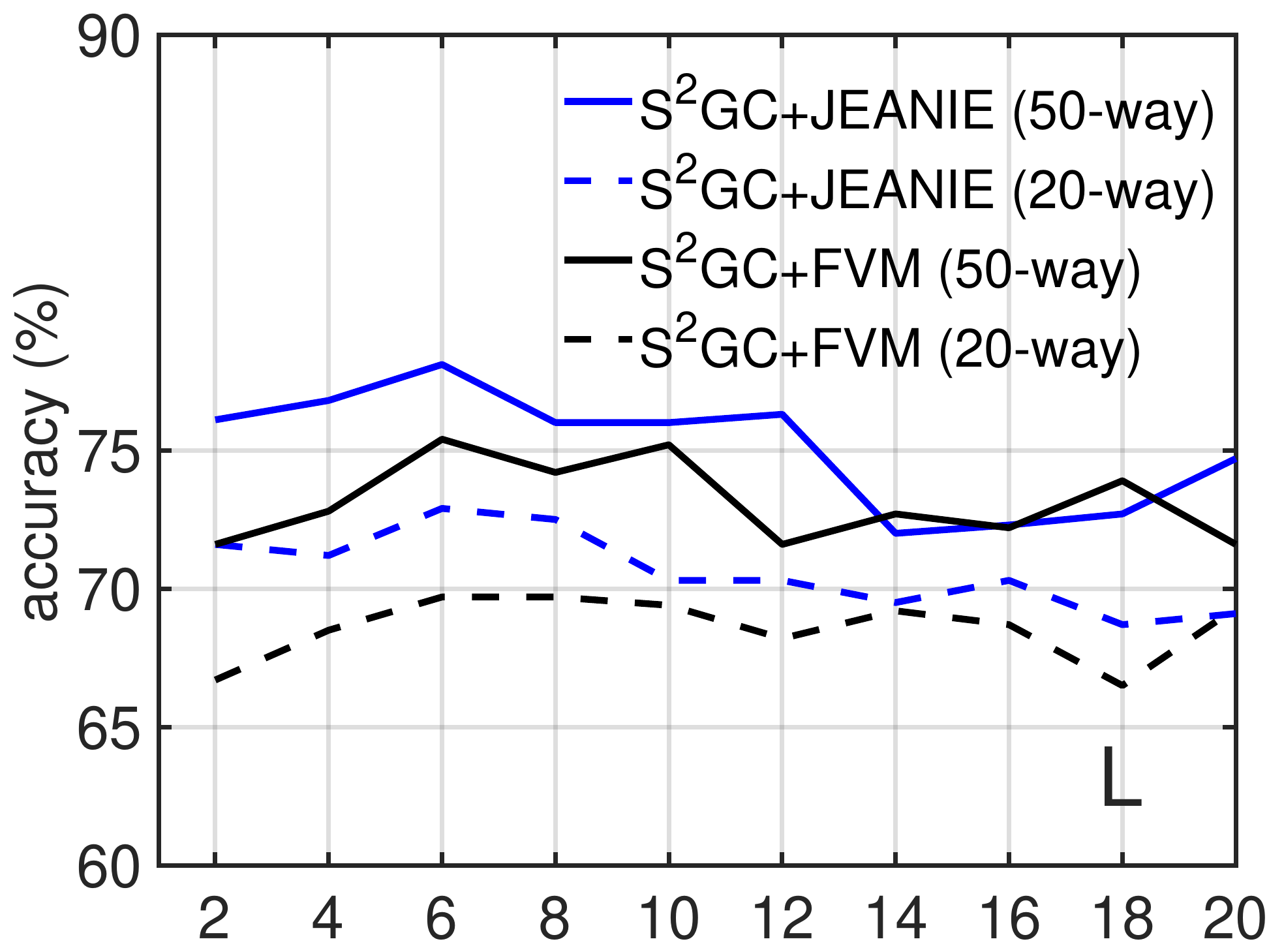}
\caption{\label{fig:ssgc_degree}}
\end{subfigure}
\caption{Evaluations of (a) $\alpha$ and (b) the number of layers $L$ for S$^2$GC on NTU-60.
}
\label{fig:ssgc_alpha_degree}
\end{figure}

Table~\ref{tab:backbone-comp} is a comparison of CNN, RNN and GNN as backbones of JEANIE. 
The role of this backbone in JEANIE is to process the per-block per-viewpoint body joint features obtained from MLP and exploit interactions among body joints. The input and output feature dimension (per temporal block) of backbone (\eg, CNN, RNN and GNN) are $J\!\times\!d$. 
For CNN, we simply use 2D convolution (square kernels and equal stride with auto padding to ensure the input and output feature maps have the same dimensions).
For RNN (joint-wise), we use $J$ RNN cells, each RNN processes $d$-dimensional vector for each body joint ($J$-input to $J$-output). Both input and output are $d$-dimensional feature vectors. The outputs from $J$ RNN cells are concatenated to form $J\!\times\!d$ feature maps. 
For RNN (temporal-wise), we modify the first MLP in our pipeline, to produce $J\!\times\!d$ vector per each time-step $1,\cdots,T$ of temporal block. We use $T$ RNN cells, each RNN processes $J\!\times\!d$-dimensional vector ($T$-input to $1$-output). The output we use is from the last of $T$ RNN cells is $J\!\times\!d$. 
As shown in the table, GNN outperforms RNN, and RNN outperforms CNN. Note that our GNN (S$^2$GC) is a simple linear projection on a spectral filter of graph and without learnable parameters (Eq. (11) of Supplementary Material). We believe RNN (temporal-wise) is better than RNN (joint-wise) as joints to not have well defined time-like order (thus RNN is bad) while GNN uses the graph connectivity (topological order).

\begin{table}[!htbp]
\caption{A comparison of different backbones in JEANIE on NTU-60 (\#training classes = 10).}
\begin{center}
\vspace{-0.1cm}
\resizebox{0.75\linewidth}{!}{\begin{tabular}{l c c c c}
\toprule
Backbones & CNN$\;\;$ & RNN (joint-wise)$\;\;$ & RNN (temporal-wise)$\;\;$ & GNN (ours) \\
\midrule
Results (acc.) & 55.3 & 59.9 & 61.1 & {\bf 65.0}\\
\bottomrule
\end{tabular}}
\label{tab:backbone-comp}
\end{center}
\end{table}

\subsection{Evaluations of viewpoint alignment}
\label{sup:vall}

Fig.~\ref{fig:gamma} shows comparisons  with temporal-viewpoint alignment ({\em V}) \vs temporal alignment alone on NTU-60. 

Fig.~\ref{fig:sgc_gcn_gamma} shows  evaluations of $\gamma$ without the viewpoint alignment. Fig.~\ref{fig:sgc_vt_gamma} shows that temporal-viewpoint alignment ({\em V}) brings around 5\% (20- and 50-way protocols, $\gamma\!=\!0.0001$) improvement. 

\subsection{Evaluations \wrt ${\alpha}$}
\label{eva_alpha}
Figure~\ref{fig:ssgc_alpha} shows the evaluations of $\alpha$ for the S$^2$GC backbone. As shown in the plot, for 50-way protocol, the best performance is achieved when $\alpha\!=\!0.7$ ($\alpha\!=\!0.5$  is the second best performer for the 50-way protocol)). For the 20-way protocol, the top performer is $\alpha\!=\!0.4$ or $\alpha\!=\!0.5$. Thus, we chose $\alpha\!=\!0.5$ in our experiments. Please note we observed the same trend on the validation split.

\subsection{Evaluations \wrt the number of layers ${L}$}
\label{eva_layer}

Figure~\ref{fig:ssgc_degree} shows the performance \wrt the number of layers $L$ used by S$^2$GC and S$^2$GC+JEANIE. As shown in this plot,  when $L\!=\!6$, S$^2$GC with JEANIE performs the best for both 20- and 50-way experiments. For the Free Viewpoint Matching (FVM) using S$^2$GC (S$^2$GC+FVM), the performance is not as stable as in the case of S$^2$GC+JEANIE. 

\subsection{Evaluation of stride for viewing angles}
\label{stride-degree}

 The stride for viewing angles is a mere equivalent of stride parameter in CNNs, which is an equal interval location sampler. We show various sampling steps for viewing angles in Table~\ref{tab:stride-degree}. We notice that when stride is 15$^\circ$, JEANIE performs the best on NTU-60.

\begin{table}[t]
\setlength{\tabcolsep}{0.25em}
\renewcommand{\arraystretch}{0.50}
\caption{Experimental results of stride for degrees on NTU-60.}
\begin{center}
\resizebox{0.36\linewidth}{!}{\begin{tabular}{ l c c c c c c}
\toprule
 & & 10 & 20  & 30  & 40  & 50 \\
\midrule
$5^\circ$ && 63.8& 73.7& 74.2& 75.0& 76.5\\
$10^\circ$& &64.2& 74.8& 75.0& 78.0& 79.1\\
\rowcolor{LightCyan}$15^\circ$&& {\bf 65.0} & {\bf 75.2}& {\bf 76.7}& {\bf 78.9}& {\bf 80.0}\\
$30^\circ$&& {\bf 65.0} & 74.8& 75.0& 76.9& 78.5\\
$45^\circ$&&60.0& 68.5& 71.0& 71.5& 72.0\\
\bottomrule
\label{tab:stride-degree}
\end{tabular}}
\end{center}
\end{table}

\section{Inference Time}

Table~\ref{tab:inference-time} below compares training and inference times per query on Titan RTX 2090. For soft-DTW, each query is augmented by $K\!\times\!K'\!=\!9$ viewpoints. In the test time, we average match distance over $K\!\times\!K'\!=\!9$ viewpoints of each test query (this is a popular standard test augmentation strategy) \wrt support samples. 
This strategy is denoted as soft-DTW$_\text{aug}$. 
We also apply the above strategy to TAP (denoted as  TAP$_\text{aug}$). 
JEANIE also uses $K\!\times\!K'\!=\!9$ viewpoints per query.
%
\begin{table}[!htbp]
\caption{A comparison of training/inference time (per query) on NTU-60 (\#training classes = 10).}
\begin{center}
\vspace{-0.1cm}
\resizebox{0.95\linewidth}{!}{
\begin{tabular}{l c c c c}
\toprule
& Training time (s)$\;\;$ & Inference time (s)$\;\;$ & Total inference time (s)$\;\;$ & Acc. (\%)\\
\midrule
soft-DTW$_\text{aug}$& 0.098 & 0.019 & 178.5 & 56.8\\
TAP$_\text{aug}$& 0.124 & 0.024 & 225.5 & 57.6\\
JEANIE &  0.099 &  0.020 & 187.0 &  65.0\\
\bottomrule
\end{tabular}}
\label{tab:inference-time}
\end{center}
\end{table}
%
We exclude the time of applying viewpoint generation as skeletons can be pre-processed at once (1.6h with non-optimized CPU code) and stored for the future use.
Among methods which use multiple viewpoints, JEANIE outperforms soft-DTW$_\text{aug}$ and TAP$_\text{aug}$ by 8.2\% and 7.4\% respectively. JEANIE outperforms ordinary soft-DTW and TAP by 11.3\% and 10.8\%. For soft-DTW$_\text{aug}$ and TAP$_\text{aug}$, their total training and testing were about $5\!\times$ and $9\!\times$ slowed compared to counterpart soft-DTW and TAP. This is expected as they had to deal with $K\!\times\!K'\!=\!9$ more samples. We tried also parallel JEANIE. Training JEANIE$_\text{par}$ with 4  Titan RTX 2090 took 44h, the total inference was 48s.

\section{More baselines on NTU-60}
\label{sup:morebase}

\begin{table}[t]
\setlength{\tabcolsep}{0.25em}
\renewcommand{\arraystretch}{0.50}
\caption{Evaluations of additional baselines on NTU-60.}
\begin{center}
\resizebox{0.95\linewidth}{!}{\begin{tabular}{ l c c c c c c }
\toprule
\# Training Classes & & 10 & 20 & 30 & 40 & 50\\ 
\midrule

Matching Nets~\citelatex{f4Matching_supp} (skeleton to image tensor)& & 26.7 & 30.6 & 32.9 & 36.4 & 39.9\\
Proto. Net~\citelatex{f1_supp} (skeleton to image tensor)&& 30.6 & 33.9 & 36.8 & 40.2 & 43.0\\
Proto. Net (per block image tensor, temp. align.)&& 40.4 & 42.4 & 45.2 & 49.0 & 50.3\\
Proto. Net (per block image tensor, temp. \& view. align.)& & 41.6 & 43.0 & 47.7 & 50.4 & 51.6\\
\bottomrule

\end{tabular}}
\label{ntu60results-other}
\end{center}
\end{table}

Table~\ref{ntu60results-other} shows more evaluations on NTU-60. Before GCNs have become mainstream backbones for the  3D Skeleton-based Action Recognition, encoding 3D body joints of skeletons as texture-like images enjoyed some limited popularity, with approaches \citelatex{supp_ke_2017_CVPR,supp_ke_tip,supp_tas2018cnnbased} feeding such images into CNN backbones. This facilitates easy FSL with existing pipelines such as Matching Nets~\citelatex{f4Matching_supp} and Prototypical Net~\citelatex{f1_supp}. Thus, we reshape the normalized 3D coordinates of each skeleton sequence or per block skeleton into image tensors, and pass them into Matching Nets~\citelatex{f4Matching_supp} and Prototypical Net~\citelatex{f1_supp} for few-shot learning. Not surprisingly, using texture-like images for skeletons is suboptimal. Our JEANIE is more than 25\% better.

\section{Drawbacks/Limitations}
Some minor drawbacks of our work are: 
(i) extending our work to video-based action recognition requires more work as it is hard to simulate the views of pixels for RGB videos; 
(ii) alignment in 4D space is slower than in 2D space and slower than no alignment at all, but improvements in accuracy are significant, and in fact the alignment step can be written as the RKHS kernel, which can be linearized by the Nystr\"om feature maps, casting 4D problem as two 2D problems.

\section{Evaluation Protocols}
\label{app:epr}

Below, we detail our new/additional evaluation protocols used in the experiments.
\subsection{Few-shot AR protocols on the small-scale datasets}
\label{small_proto}
Below, we explain the selection process.

\paragraph{FSAR (MSRAction3D)}. As this dataset contains 20 action classes, we randomly choose 10 action classes for training and the rest 10 for testing. We repeat this sampling process 10 times to form in total 10 train/test splits. For each split, we have 5-way and 10-way experimental settings. The overall performance on this dataset is computed by averaging the performance on 10 splits.

\paragraph{FSAR (3D Action Pairs)}. This dataset has in total 6 action pairs (12 action classes), each pair of action has very similar motion trajectories, \eg, {\it pick up a box} and {\it put down a box}. We randomly choose 3 action pairs to form a training set (6 action classes) and the half action pairs for the test set, and in total there are ${\binom nk}\!=\!{\binom {6}{3}\!=\!20}$ different combinations of train/test splits. As our train/test splits are based on action pairs, we are able to test  whether the algorithm is able to classify unseen action pairs that share similar motion trajectories. We use 5-way protocol on this dataset to evaluate the performance of FSAR, averaged over all 20 splits.

\paragraph{FSAR (UWA3D Activity)}. This dataset has 30 action classes. We randomly choose 15 action classes for training and the rest half action classes for testing. We form in total 10 train/test splits, and we use 5-way and 10-way protocols on this dataset, averaged over all 10  splits.

\subsection{One-shot protocol on NTU-60}

Following NTU-120~\citelatex{Liu_2019_NTURGBD120_supp}, we introduce the one-shot AR setting on NTU-60. We split the whole dataset into two parts: auxiliary set (on NTU-120 the training set is called as auxiliary set, so we follow such a terminology) and one-shot evaluation set. 

\paragraph{Auxiliary set} contains 50 classes, and all samples of these classes can be used for learning and validation. Evaluation set consists of 10 novel classes, and one sample from each novel class is picked as the exemplar (terminology introduced by authors of NTU-120), while all the remaining samples of these classes are used to test the recognition performance. 

\paragraph{Evaluation set} contains 10 novel classes, namely, A1, A7, A13, A19, A25, A31, A37, A43, A49, A55. The following 10 samples are the exemplars:

\noindent(01)S001C003P008R001A001, (02)S001C003P008R001A007,\\ (03)S001C003P008R001A013, (04)S001C003P008R001A019, \\(05)S001C003P008R001A025, (06)S001C003P008R001A031, \\(07)S001C003P008R001A037, (08)S001C003P008R001A043,\\ (09)S001C003P008R001A049, (10)S001C003P008R001A055. 


\paragraph{Auxiliary set} contains  50 classes (the remaining 50 classes of NTU-60 excluding the 10 classes in evaluation set).

\subsection{Few-shot multiview classification on NTU-120}

\paragraph{Horizontal camera view}.
As NTU-120 is captured by 3 cameras (from 3 different horizontal angles: -45$^{\circ}$, 0$^{\circ}$, 45$^{\circ}$), we split the whole dataset based on the camera ID to form our 3 horizontal camera viewpoints (left, center and right views). We then evaluate few-shot multiview classification using (i) the left view for training and the center view for testing (ii) the left view for training and the right view for testing (ii) the left and center views for training and the right view for testing.

\paragraph{Vertical camera view}. Based on the table provided in~\citelatex{Liu_2019_NTURGBD120_supp}, we first group 32 camera setups into 3 groups by dividing the range of heights into 3 equally-sized ranges to form roughly the top, center and bottom views. We then group the whole dataset into 3 camera viewpoints based on the camera setup IDs. For few-shot multiview classification, we evaluate our proposed method using (i) bottom view for training and center view for testing (ii) bottom view for training and top view for testing (iii) bottom and center views for training and top view for testing.

\section{Network configuration and training details}
\label{network_train}

Below we provide the details of network configuration and training process in the following sections.

\subsection{Network configuration}

Given the temporal block size $M$ (the number of frames in a block) and desired output size $d$, the configuration of the 3-layer MLP unit is: FC ($3M \rightarrow 6M$), LayerNorm (LN) as in \citelatex{dosovitskiy2020image_supp}, ReLU, FC ($6M \rightarrow 9M$), LN, ReLU, Dropout (for smaller datasets, the dropout rate is 0.5; for large-scale datasets, the dropout rate is 0.1), FC ($9M \rightarrow d$), LN. Note that $M$ is the temporal block size
~and $d$ is the output feature dimension per body joint. Note that ablations on the value of $M$ are already conducted in Table \ref{blockframe_overlap}.

\paragraph{Backbone with GNN and Transformer}. 
Following EN described in Section \ref{sec:appr}, let us take the query input $\mX\!\in\!\mbr{3\times J\times M}$ for the temporal block of length $M$ as an example, where $3$ indicates 3D Cartesian coordinate and $J$ is the number of body joints. As alluded to earlier, we obtain $\widehat{\mX}^T\!=\!\text{MLP}(\mX; \mathcal{F}_{MLP})\!\in\!\mbr{{d}\times J}$.

Subsequently, we employ a GNN and the transformer encoder \citelatex{dosovitskiy2020image_supp} which consists of alternating layers of Multi-Head Self-Attention (MHSA) and a feed-forward MLP (two FC layers with a GELU non-linearity between them). LayerNorm (LN) is applied before every block, and residual connections after every block. Each block feature matrix $\widehat{\mX} \in \mathbb{R}^{J \times {d}}$ encoded by GNN (without learnable $\bf \Theta$) is then passed to the transformer. Similarly to the standard transformer, we prepend a learnable vector  
${\bf y}_\text{token}\!\in\!\mbr{1\times {d}}$
to the sequence of block features $\widehat{\mX}$ obtained from GNN,
and we also add the positional embeddings ${\bf E}_\text{pos} \in \mathbb{R}^{(1+J) \times {d}}$ based on the sine and cosine functions (standard in transformers) so that token ${\bf y}_\text{token}$ and each body joint enjoy their own unique positional encoding. 
We obtain  $\mZ_0\!\in\!\mbr{(1+J)\times {d}}$ which is the input in the following backbone:
\begin{align}
&{\bf Z}_0 = [{\bf y}_\text{token}; \text{GNN}(\widehat{\mX})]+{\bf E}_\text{pos}, \label{eq:proj}\\
&{\bf Z}^\prime_k = \text{MHSA}(\text{LN}({\bf Z}_{k-1})) + {\bf Z}_{k-1}, \;k = 1, \cdots, L_\text{tr}\label{eq:mhsa}\\
& {\bf Z}_k = \text{MLP}(\text{LN}({\bf Z}^\prime_k)) + {\bf Z}^\prime_k, \qquad\quad \,k = 1, \cdots, L_\text{tr}\label{eq:mlp} \\
& \vy' = \text{LN}\big(\mZ^{(0)}_{L_\text{tr}}\big) \qquad\qquad\text{ where }\quad\;\; \vy'\in \mathbb{R}^{1 \times {d}} \label{eq:blockfeat}\\
&f(\mX; \mathcal{F})=\text{FC}(\vy'^T; \mathcal{F}_{FC})\qquad\qquad\quad\!\in \mathbb{R}^{d'},\label{eq:final_fc_bl}
\end{align}
where $\mZ^{(0)}_{L_\text{tr}}$ is the first ${d}$ dimensional row vector extracted from the output matrix $\mZ_{L_\text{tr}}$  of size $(J\!+\!1)\times{d}$ which corresponds to the last layer $L_\text{tr}$ of the transformer. Moreover, parameter $L_\text{tr}$ controls the depth of the transformer, whereas $\mathcal{F}\!\equiv\![\mathcal{F}_{MLP},\mathcal{F}_{GNN},\mathcal{F}_{Transf},\mathcal{F}_{FC}]$ is the  set of  parameters of EN. In case of APPNP, SGC and S$^2$GC, $|\mathcal{F}_{GNN}|\!=\!0$ because we do not use their learnable parameters $\bf \Theta$ (\ie, think $\bf \Theta$ is set as the identity matrix).

As in Section \ref{sec:appr}, one can define now a support feature map as $\mPsi'\!=\![f(\boldsymbol{X}_1;\mathcal{F}),\cdots,f(\boldsymbol{X}_{\tau'};\mathcal{F})]\!\in\!\mbr{d'\times\tau'}$ for $\tau'$ temporal blocks, and the query map $\mPsi$ accordingly.

 The hidden size of our transformer  (the output size of the first FC layer of the MLP in Eq. \eqref{eq:mlp}) depends on the dataset. For smaller datasets, the depth of the transformer is $L_\text{tr}\!=\!6$ with $64$ as the hidden size, and the MLP output size is  ${d}\!=\!32$ (note that the MLP which provides $\widehat{\mX}$ and the MLP in the transformer must both have  the same output size). For NTU-60, the depth of the transformer is $L_\text{tr}\!=\!6$, the hidden size is 128 and the MLP output size is  ${d}\!=\!64$. For NTU-120, the depth of the transformer is   $L_\text{tr}\!=\!6$, the hidden size is 256 and the MLP size is ${d}\!=\!128$. For Kinetics-skeleton, the depth for the transformer is $L_\text{tr}\!=\!12$, hidden size is 512 and the MLP output size is ${d}\!=\!256$. The number of Heads for the transformer of smaller datasets, NTU-60, NTU-120 and Kinetics-skeleton is set as 6, 12, 12 and 12, respectively.

The output sizes $d'$ of the final FC layer in Eq. \eqref{eq:final_fc_bl} are 50, 100, 200, and 500 for the smaller datasets, NTU-60, NTU-120 and Kinetics-skeleton, respectively.

\subsection{Training details}
The weights for the pipeline are initialized with the normal distr. (zero mean and unit standard dev.).
We use 1e-3 for the learning rate, and the weight decay is 1e-6. We use the SGD optimizer.
We set the number of training episodes to 100K for NTU-60, 200K for NTU-120, 500K for 3D Kinetics-skeleton, 10K for small datasets such as UWA3D Multiview Activity II.                        
We use Hyperopt for hyperparam. search on validation sets for all the datasets.

\section{Skeleton Data Preprocessing}

Before passing the skeleton sequences into MLP and graph networks (\eg, S$^2$GC), we first normalize each body joint \wrt to the torso joint ${\bf v}_{f, c}$:
\begin{equation}
    {\bf v}^\prime_{f, i}\!=\!{\bf v}_{f, i}\!-\!{\bf v}_{f, c},
\end{equation}
where $f$ and $i$ are the index of video frame and human body joint respectively. After that, we further normalize each joint coordinate into  [-1, 1] range:

\begin{equation}
    \hat{{\bf v}}_{f, i}[j] = \frac{{\bf v}^\prime_{f, i}[j]}{ \text{max}([\text{abs}({\bf v}^\prime_{f, i}[j])]_{f\in\idx{\tau},i\in\idx{J} } )},
\end{equation}
where $j$ is for selection of the $x$, $y$ and $z$ axes, $\tau$ is the number of frames and $J$ is the number of 3D body joints per frame.

For the skeleton sequences that have more than one performing subject, (i) we normalize each skeleton separately, and each skeleton is passed to MLP for learning the temporal dynamics, and (ii) for the output features per skeleton from MLP, we pass them separately to graph networks, \eg, two skeletons from a given video sequence will have two outputs from the graph networks, and we aggregate the outputs through average pooling before passing to FVM or JEANIE.

\section{Additional Visualizations}
\label{supp:visualization}

To explain what makes JEANIE perform well on the task of comparing pairs of sequences, we perform additional visualisations.

To this end, we choose skeleton sequences from UWA3D Multiview Activity II for experiments and visualizations of FVM and JEANIE. UWA3D Multiview Activity II  contains rich viewpoint configurations and so is perfect for our investigations. 

\noindent{\bf 1. Matching similar actions.} We choose a {\it walking} skeleton sequence (`{\tt a12\_s01\_e01\_v01}') as the query sample with more viewing angles for the camera viewpoint simulation, and we select another {\it walking} skeleton sequence of a different view (`{\tt a12\_s01\_e01\_v03}') and a {\it running} skeleton sequence (`{\tt a20\_s01\_e01\_v02}') as support samples respectively, to verify that our JEANIE is able to find the better matching distances compared to FVM.

\noindent{\bf 2. Matching actions with similar motion trajectories.} We choose a {\it two hand punching} skeleton sequence (`{\tt a04\_s01\_e01\_v01}') as the query sample with more viewing angles for the camera viewpoint simulation, and we select another {\it two hand punching} skeleton sequence of a different view (`{\tt a04\_s05\_e01\_v02}') and a {\it holding head} skeleton sequence (`{\tt a10\_s05\_e01\_v02}') as support samples respectively, to verify that our JEANIE is able to find the better matching distances compared to FVM.

\begin{figure}[!htbp]
\centering
\begin{subfigure}[b]{0.65\linewidth}
\includegraphics[trim=1.5cm 2.4cm 1.5cm 3.5cm, clip=true,width=0.99\linewidth]{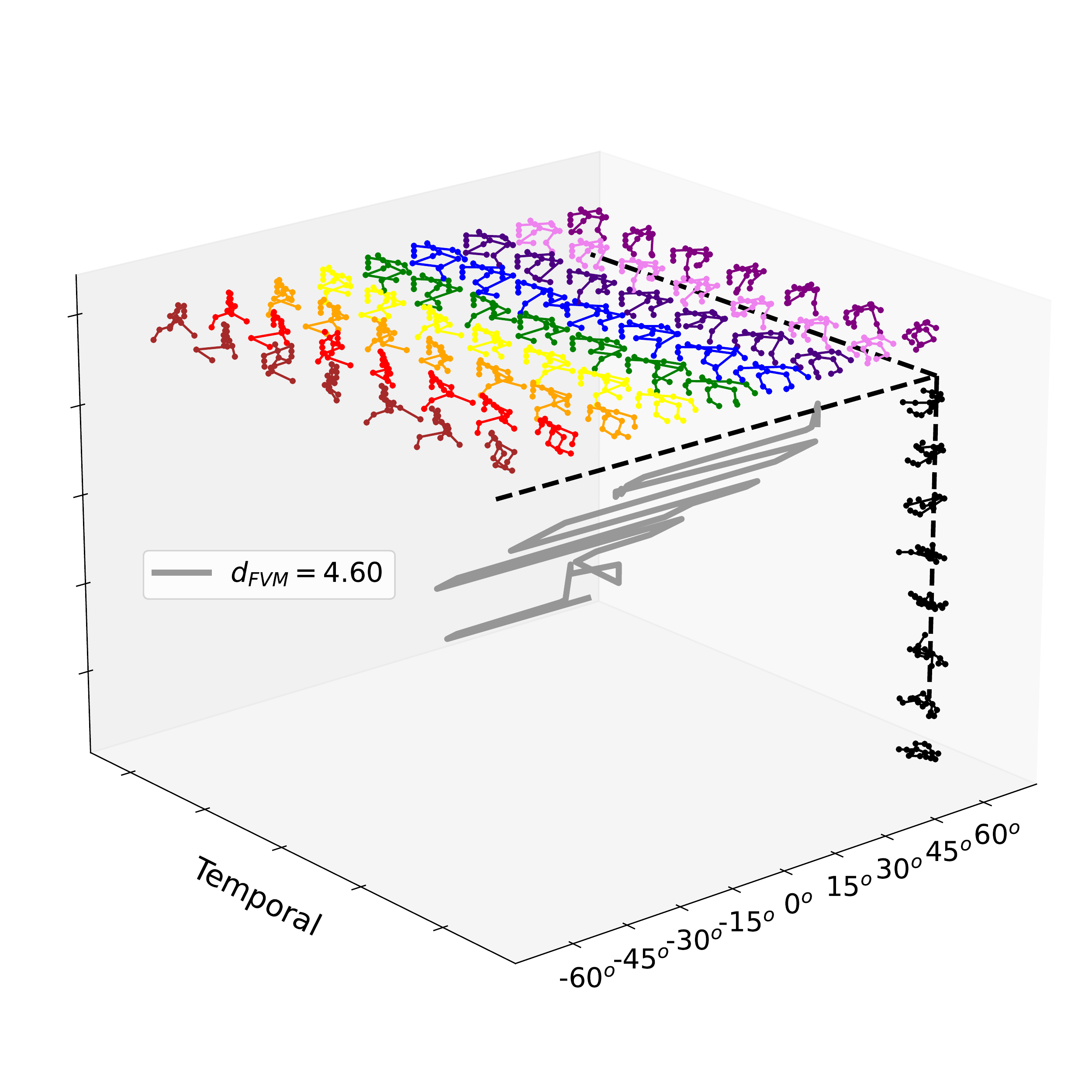}
\caption{{\it walking} \vs~{\it walking} ($d_\text{FVM}\!=\!4.60$)}\label{fig:fvm_walk_walk}
\end{subfigure}
\begin{subfigure}[b]{0.65\linewidth}
\includegraphics[trim=1.5cm 2.4cm 1.5cm 3.5cm, clip=true,width=0.99\linewidth]{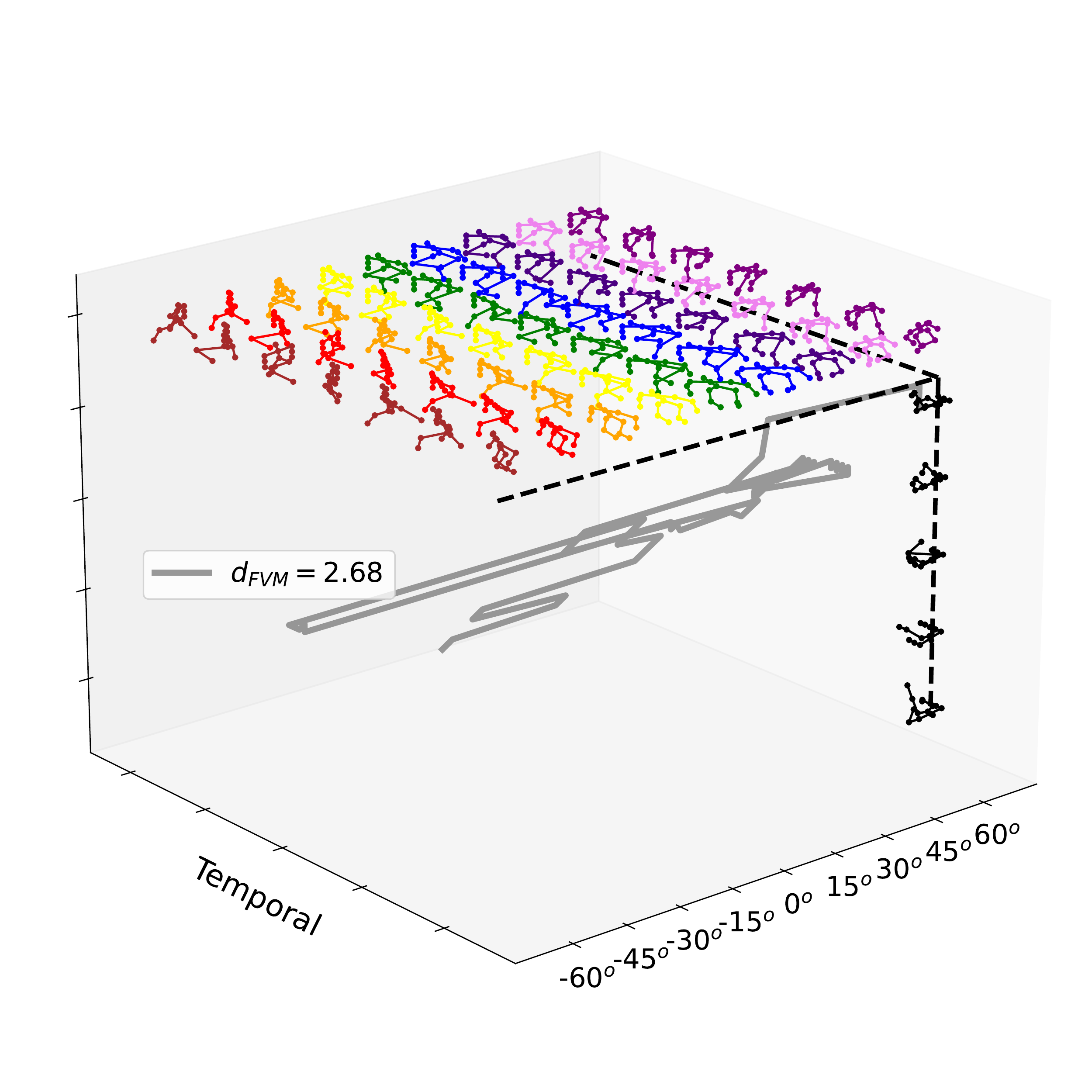}
\caption{{\it walking} \vs~{\it running} ($d_\text{FVM}\!=\!2.68$)}\label{fig:fvm_walk_run}
\end{subfigure}
\caption{Visualization of FVM for {\it walking} \vs~{\it walking} (two different sequences) and {\it walking} \vs~{\it running}. From UWA3D Multiview Activity II, we choose a {\it walking} sequence as the query sample (`{\tt a12\_s01\_e01\_v01}'). We choose another {\it walking} sequence from a different view (`{\tt a12\_s01\_e01\_v03}') and a {\it running} sequence (`{\tt a20\_s01\_e01\_v02}') as the support samples respectively. We notice that for two different action sequences in (b), the greedy FVM finds the path with a very small distance $d_\text{FVM}\!=\!2.68$ but for sequences of the same action class, FVM gives $d_\text{FVM}\!=\!4.60$. This is clearly suboptimal as the within-class distance is higher then the between-class distance (to counteract this issue, we have proposed JEANIE).}
\label{fig:fvm_vs_jeanie_supp}
\end{figure}

\begin{figure}[!htbp]
\centering
\begin{subfigure}[b]{0.65\linewidth}
\includegraphics[trim=1.5cm 2.4cm 1.5cm 3.5cm, clip=true,width=0.99\linewidth]{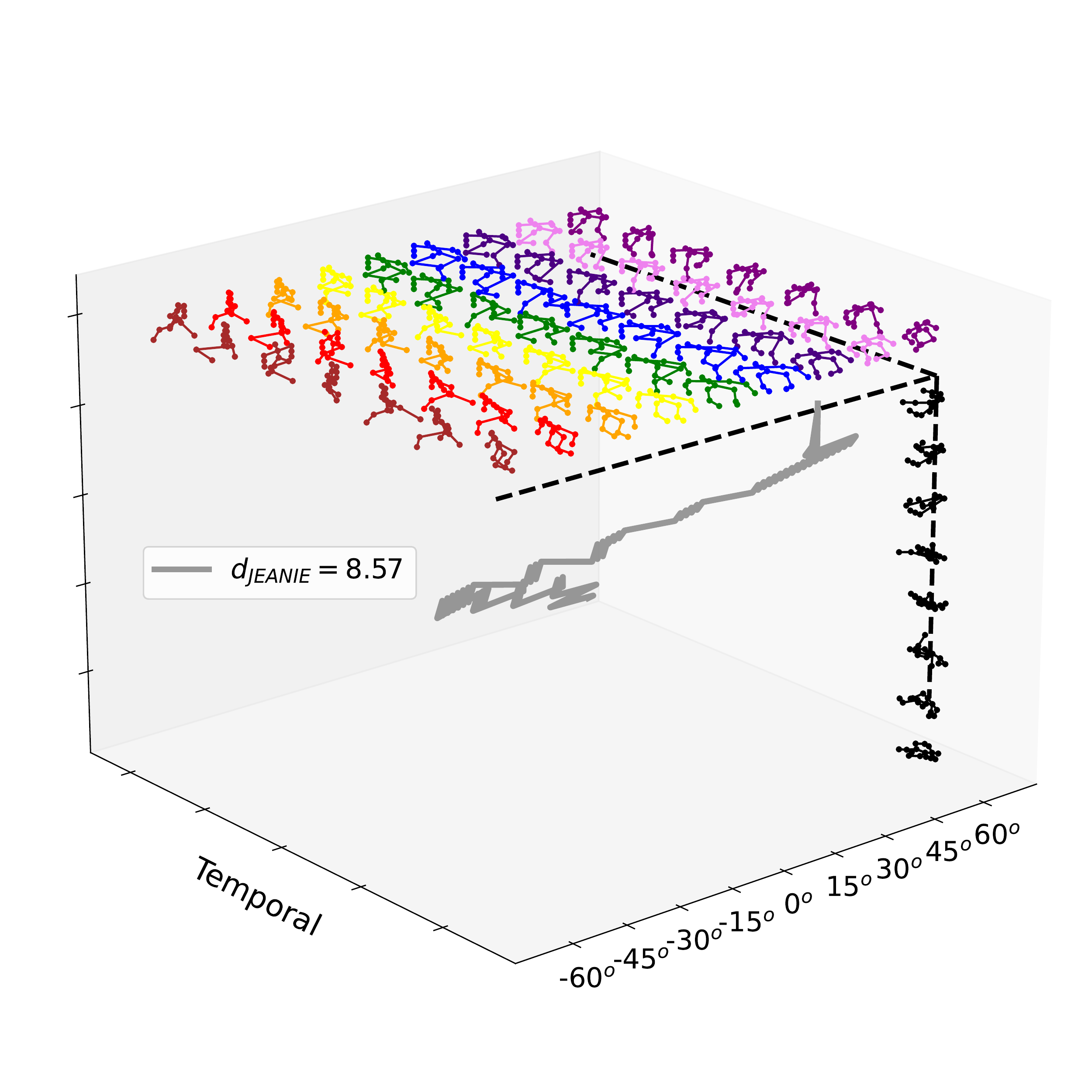}
\caption{{\it walking} \vs~{\it walking} ($d_\text{JEANIE}\!=\!8.57$)}\label{fig:jeanie_walk_walk}
\end{subfigure}
\begin{subfigure}[b]{0.65\linewidth}
\includegraphics[trim=1.5cm 2.4cm 1.5cm 3.5cm, clip=true,width=0.99\linewidth]{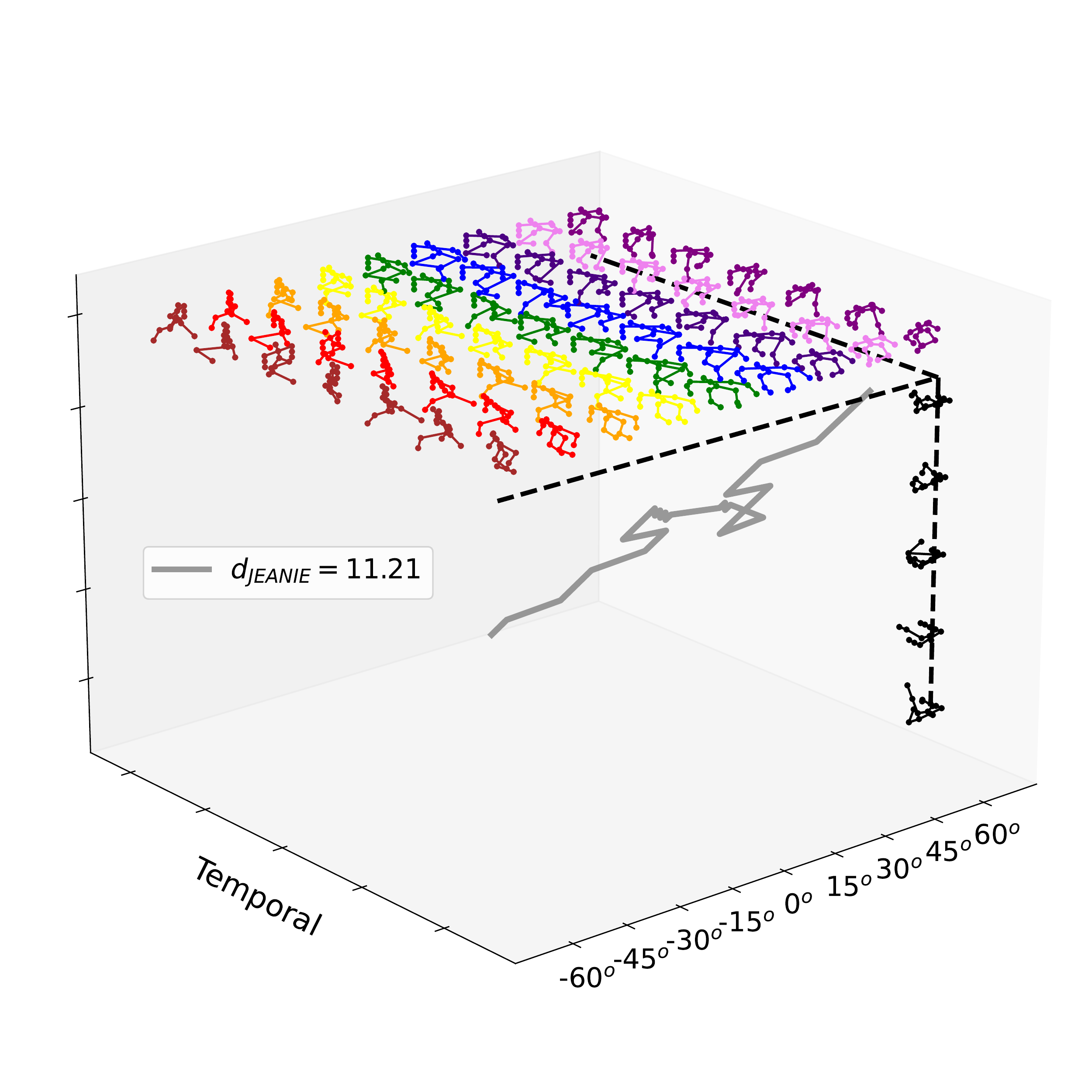}
\caption{{\it walking} \vs~{\it running} ($d_\text{JEANIE}\!=\!11.21$)}\label{fig:jeanie_walk_run}
\end{subfigure}
\caption{Visualization of JEANIE for {\it walking} \vs~{\it walking} (two different sequences) and {\it walking} \vs~{\it running}. From UWA3D Multiview Activity II, we choose a {\it walking} sequence as the query sample (`{\tt a12\_s01\_e01\_v01}'). We also choose  another {\it walking} sequence from a different view (`{\tt a12\_s01\_e01\_v03}') and a {\it running} sequence (`{\tt a20\_s01\_e01\_v02}') as the support samples respectively. In contrast to FVM in Fig. \ref{fig:fvm_vs_jeanie_supp}, our JEANIE is able to produce a smaller distance for within-class sequences and a larger distance for between-class sequences, which is a very important property when comparing pairs of sequences.}
\label{fig:fvm_vs_jeanie_supp2}
\end{figure}

Fig.~\ref{fig:fvm_vs_jeanie_supp} and~\ref{fig:fvm_vs_jeanie_supp2} show the visualizations. Comparing Fig.~\ref{fig:fvm_walk_walk} and~\ref{fig:fvm_walk_run} of FVM, we notice that for skeleton sequences from different action classes ({\it walking} \vs {\it running}), FVM finds the path with a very small distance $d_\text{FVM}\!=\!2.68$. In contrast, for sequences from the same action class ({\it walking} \vs {\it walking}), FVM gives $d_\text{FVM}\!=\!4.60$ which higher than in case of within-class sequences. This is an undesired effect which may result in wrong comparison decision. 

In contrast, in Fig~\ref{fig:jeanie_walk_walk} and~\ref{fig:jeanie_walk_run}, our JEANIE gives $d_\text{JEANIE}\!=\!8.57$ for sequences of the same action class and $d_\text{JEANIE}\!=\!11.21$ for sequences from different action classes, which means that the within-class distances are smaller than between-class distances. This is in fact a very important property when comparing pairs of sequences.

\begin{figure}[!htbp]
\centering
\begin{subfigure}[b]{0.65\linewidth}
\includegraphics[trim=1.5cm 2.4cm 1.5cm 3.5cm, clip=true,width=0.99\linewidth]{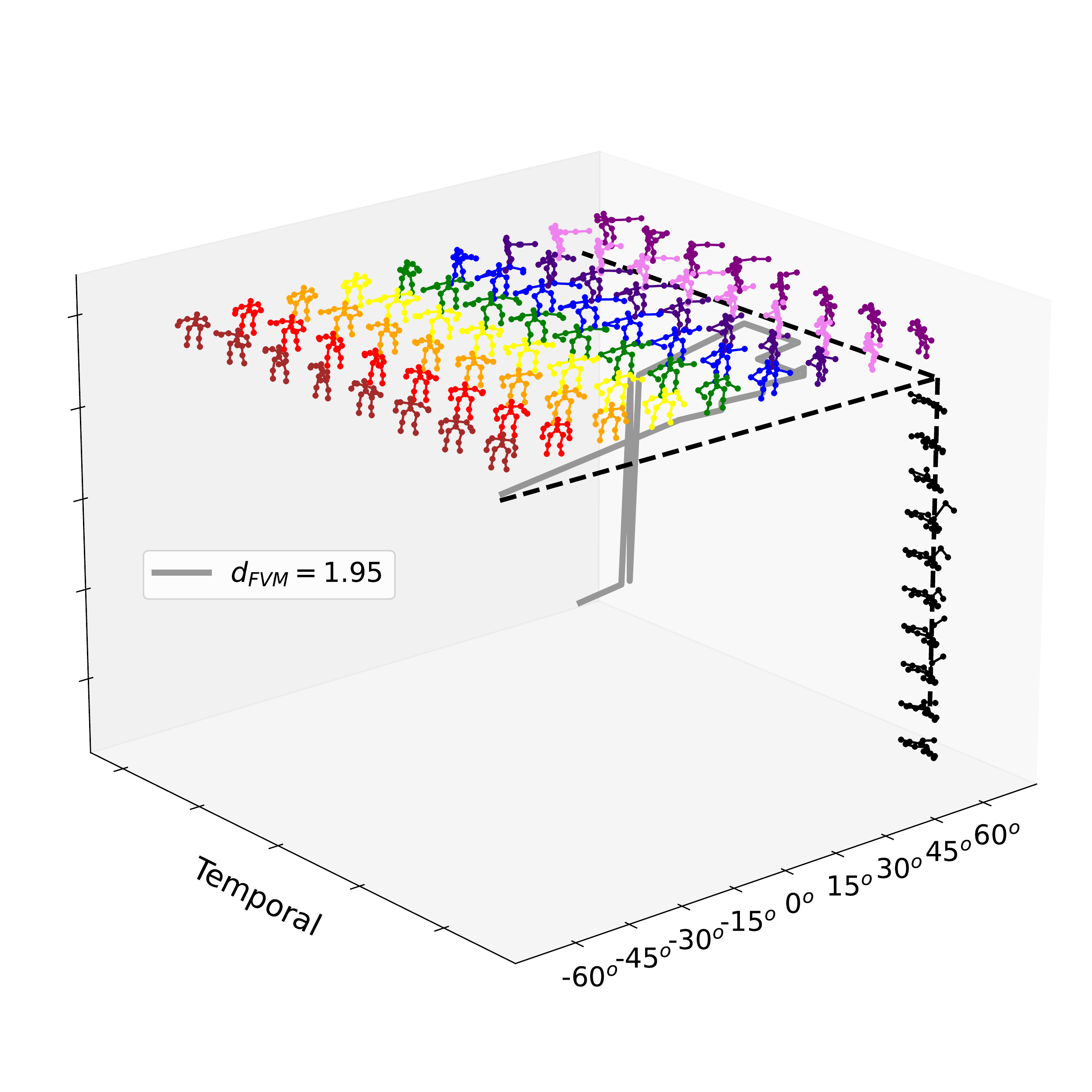}
\caption{{\it two hand punching} \vs~{\it two hand punching} ($d_\text{FVM}\!=\!1.95$)}\label{fig:fvm_punch_punch}
\end{subfigure}
\begin{subfigure}[b]{0.65\linewidth}
\includegraphics[trim=1.5cm 2.4cm 1.5cm 3.5cm, clip=true,width=0.99\linewidth]{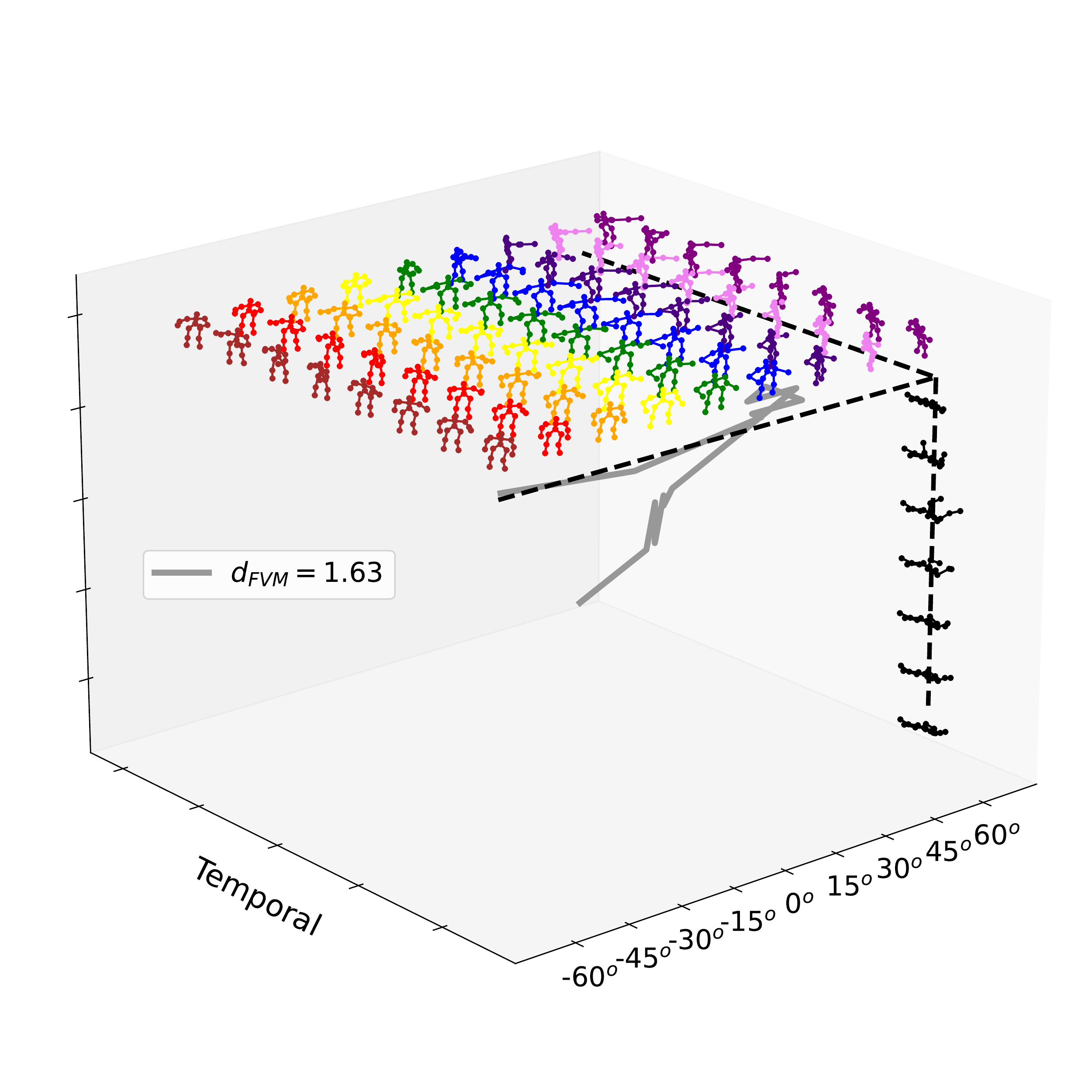}
\caption{{\it two hand punching} \vs~{\it holding head} ($d_\text{FVM}\!=\!1.63$)}\label{fig:fvm_punch_hold}
\end{subfigure}
\caption{Visualization of FVM for {\it two hand punching} \vs~{\it two hand punching} (two different sequences) and {\it two hand punching} \vs~{\it holding head}. From UWA3D Multiview Activity II, we choose a {\it two hand punching} sequence as the query sample (`{\tt a04\_s01\_e01\_v01}'), and another {\it two hand punching} sequence from a different view (`{\tt a04\_s05\_e01\_v02}') and a {\it holding head} sequence (`{\tt a10\_s05\_e01\_v02}') as the support samples respectively. We notice that for two different action sequences in (b), the greedy FVM finds the path which results in $d_\text{FVM}\!=\!1.63$ for sequences of different action classes, yet FVM gives $d_\text{FVM}\!=\!1.95$ for two sequences of the same class. The within-class distance should be smaller than the between-class distance but greedy approaches such as FVM cannot handle this requirement well.}
\label{fig:fvm_vs_jeanie_supp3}
\end{figure}

\begin{figure}[!htbp]
\centering
\begin{subfigure}[b]{0.65\linewidth}
\includegraphics[trim=1.5cm 2.4cm 1.5cm 3.5cm, clip=true,width=0.99\linewidth]{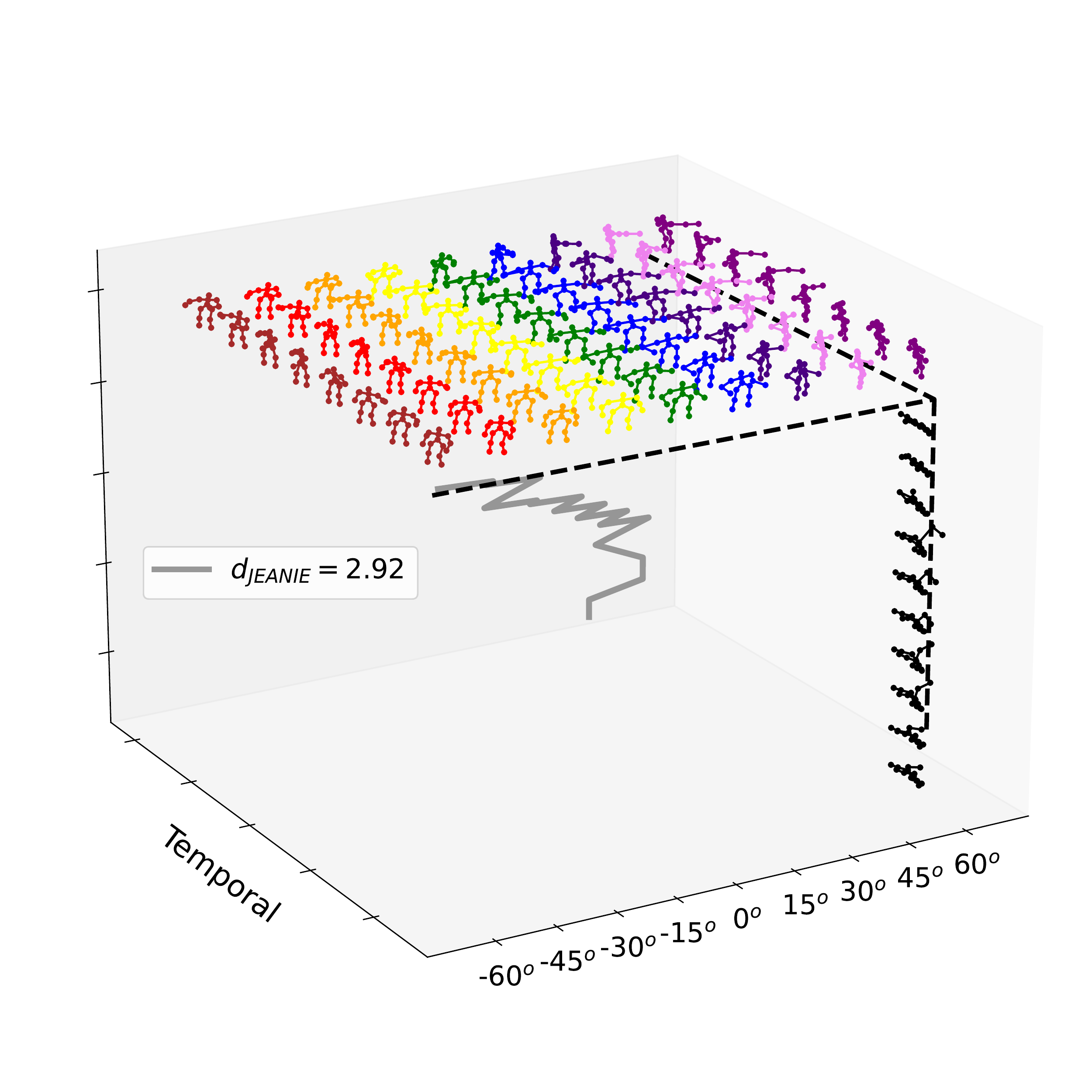}
\caption{{\it two hand punching} \vs~{\it two hand punching} ($d_\text{JEANIE}\!=\!2.92$)}\label{fig:jeanie_punch_punch}
\end{subfigure}
\begin{subfigure}[b]{0.65\linewidth}
\includegraphics[trim=1.5cm 2.4cm 1.5cm 3.5cm, clip=true,width=0.99\linewidth]{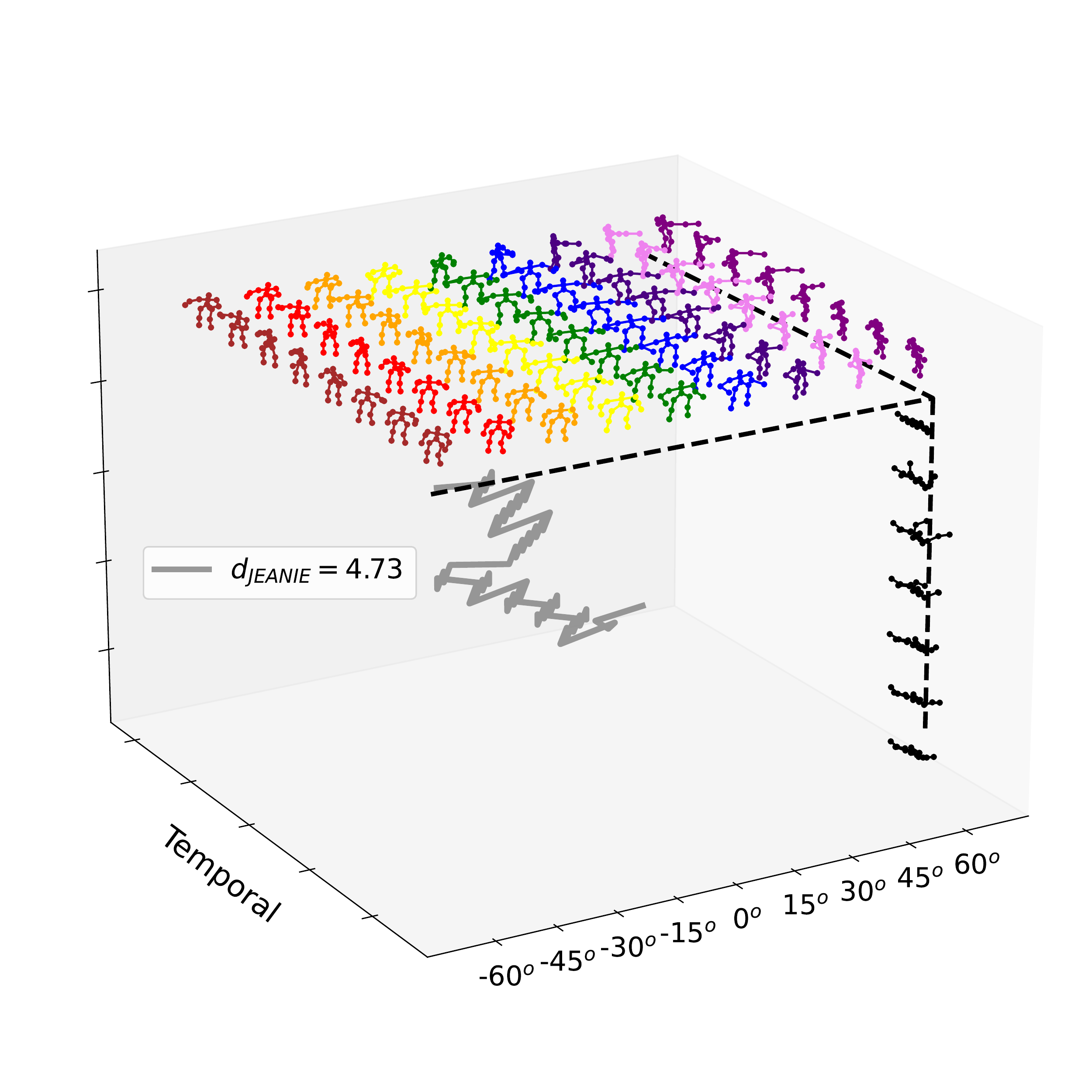}
\caption{{\it two hand punching} \vs~{\it holding head} ($d_\text{JEANIE}\!=\!4.73$)}\label{fig:jeanie_punch_hold}
\end{subfigure}
\caption{Visualization of JEANIE for {\it two hand punching} \vs~{\it two hand punching} (two different sequences) and {\it two hand punching} \vs~{\it holding head}.  From UWA3D Multiview Activity II, we choose a {\it two hand punching} sequence as the query sample (`{\tt a04\_s01\_e01\_v01}'), and another {\it two hand punching} sequence from a different view (`{\tt a04\_s05\_e01\_v02}') and a {\it holding head} sequence (`{\tt a10\_s05\_e01\_v02}') as the support samples respectively. Our JEANIE gives smaller distance when comparing within-class sequences compared to between-class sequences. This is a very important property when comparing pairs of sequences.}
\label{fig:fvm_vs_jeanie_supp4}
\end{figure}

Fig~\ref{fig:fvm_vs_jeanie_supp3} and~\ref{fig:fvm_vs_jeanie_supp4} show additional visualizations. Again, our JEANIE produces more reasonable matching distances than FVM.

\bibliographystylelatex{splncs04}
\bibliographylatex{egbib}

\end{document}